\documentclass{article}

\usepackage[nonatbib, final]{nips_2017}

\usepackage[utf8]{inputenc} 
\usepackage[T1]{fontenc}    
\usepackage{hyperref}       
\usepackage{url}            
\usepackage{booktabs}       
\usepackage{amsfonts}       
\usepackage{nicefrac}       
\usepackage{microtype}      
\usepackage{cite}
\usepackage{graphicx}
\usepackage{subcaption}
\usepackage{pdfpages}
\pdfoutput=1

\title{Ablation Studies in Artificial Neural Networks}

\author{
  Richard Meyes$^{1,2}$, Melanie Lu$^{2}$, Constantin Waubert de Puiseau$^{2}$, Tobias Meisen{$^{1,2}$} \vspace{6pt} \\
  $^{1}$Institute of Technologies and Management of the Digital Transformation, \\
  Bergische Universität Wupeprtal, Rainer-Gruenter-Straße 21, 42119 Wuppertal, Germany \vspace{2pt} \\
    $^{2}$Institute of Information Management in Mechanical Engineering, \\
  RWTH Aachen University, Dennewartstr. 27, 52064 Aachen, Germany \vspace{6pt} \\
  \texttt{\{richard.meyes, melanie.lu, constantin.waubert, } \\
  \texttt{tobias.meisen\}@ima-ifu.rwth-aachen.de} \\
}

\begin{document}

\maketitle

\begin{abstract}
Ablation studies have been widely used in the field of neuroscience to tackle complex biological systems such as the extensively studied Drosophila central nervous system, the vertebrate brain and more interestingly and most delicately, the human brain. In the past, these kinds of studies were utilized to uncover structure and organization in the brain, i.e. a mapping of features inherent to external stimuli onto different areas of the neocortex. considering the growth in size and complexity of state-of-the-art artificial neural networks (ANNs) and the corresponding growth in complexity of the tasks that are tackled by these networks, the question arises whether ablation studies may be used to investigate these networks for a similar organization of their inner representations. In this paper, we address this question and performed two ablation studies in two fundamentally different ANNs to investigate their inner representations of two well-known benchmark datasets from the computer vision domain. We found that features distinct to the local and global structure of the data are selectively represented in specific parts of the network. Furthermore, some of these representations are redundant, awarding the network a certain robustness to structural damages. We further determined the importance of specific parts of the network for the classification task solely based on the weight structure of single units. Finally, we examined the ability of damaged networks to recover from the consequences of ablations by means of recovery training. We argue that ablations studies are a feasible method to investigate knowledge representations in ANNs and are especially helpful to examine a networks robustness to structural damages, a feature of ANNs that will become increasingly important for future safety-critical applications. Our code is publicly available  
\footnote{https://github.com/RichardMeyes/AblationStudies} to reproduce our results and build upon them.
\end{abstract}
\section{Introduction}
Recent research on deep learning (DL) has brought forth a number of remarkable applications for different problems in a variety of domains. Examples are visual object recognition, object detection and semantic segmentation in the field of computer vision (CV) \cite{krizhevsky2012imagenet, simonyan2014very, szegedy2015going, he2016deep, girshick2014rich}, speech recognition and speech separation in the field of natural language processing (NLP) \cite{hinton2012deep, graves2013speech, young2018recent, chen2018dnn, luo2018speaker} or self-learning agents based on deep reinforcement learning (DRL) for video games \cite{mnih2013playing, mnih2015human, hessel2017rainbow, pohlen2018observe}, classic board games \cite{tesauro1995temporal, silver2016mastering, silver2017mastering} as well as locomotion and robotic control \cite{lillicrap2015continuous, levine2016end, schulman2017proximal, tobin2017domain, andrychowicz2017hindsight, andrychowicz2018learning}. During the last few years, the strong increase in availability of computational resources combined with the facilitation of new computing paradigms such as GPU programming \cite{krizhevsky2012imagenet} and asynchronous methods for training deep neural networks (DNNs) \cite{mnih2016asynchronous, espeholt2018impala} resulted in an increase of the average size, i.e. the number of trainable weights, of state-of-the-art DNNs. These networks exhibit holistic behavior which cannot simply be explained by considering only the functional mechanisms of key components of the network, such as single units, their activation functions, regularization mechanisms, etc. Despite this development, the main research focus in the past was placed on increasing the performance and speed of these networks solving specific benchmark tasks rather than on the development of new methods and perspectives to understand how knowledge, which is acquired during training, is represented in these complex networks. Considering that the research on DNNs has only recently been confronted with larger networks, methods and perspectives from the field of neuroscience, a research field, which has been dealing with large and complex neural systems for decades, may prove useful to investigate the structure of knowledge representation in state-of-the-art DNNs.

In this paper, we follow a neuroscience-inspired approach based on the idea of ablation studies to analyze the structure of knowledge represented in DNNs. In these studies, neural tissue is damaged in a controlled manner while investigating how the inflicted damage influences the brain's capabilities to perform a specific task. This way, insights about the functional role of the damaged brain regions as well as insights about the structure and organization of processing external stimuli in the brain can be gained. One of the most prominent examples for such an organized structure is the cortical homunculus found in the primary motor cortex and the primary sensory cortex of primates and humans. The homunculus is a distorted representation of the human body mapped onto specific regions of the neocortex responsible for processing motor or sensory functions for different parts of the body. In the past, ablation studies were used to uncover structure and organization in other parts of the brain than the motor-cortex. For instance, neonatal cochlear ablations in cats revealed that binaural interactions, i.e. the perception of sound via intensity differences arriving at the two ears, are exhibited early in postnatal life, well before structural maturation of the auditory pathways from the ear to the cortex is complete \cite{reale1987maps}. In another study, the ablation of subplate neurons in the visual cortex of adult cats revealed their role for the functional development of ocular dominance \cite{kanold2003role}. Considering that ablation studies proved to be a valuable method to investigate large, complex neural systems, like the brain of vertebrates and primates, it seems reasonable to investigate their potential for tackling state-of-the-art artificial neural systems.

Specifically, we performed ablation studies in two ANNs. Conducting single unit ablations in a small, shallow multi layer perceptron (MLP) we investigated correlations between the spatial as well as structural characteristics of the units and their contribution to the overall performance as well as the class-specific performance of the network. We found, that some single units are important for the overall classification performance, while other single units are only selectively important for a specific class. Furthermore, the importance of a single unit for the classification performance correlates with the extent to which the unit's weight distribution of incoming connections after training differs from the initial randomly initialized distribution. We further investigated the robustness of the classification performance by checking for redundant knowledge representations of specific classes in different areas of the network utilizing pairwise unit ablations. The results showed that pairwise unit ablations have a stronger effect on the classification performance than the summed effects of single ablations of the same units. Second, we investigated a larger state-of-the-art convolutional neural network (CNN) for correlations between the size as well as the depth of the ablated portions of the network and the overall performance as well as the class-specific performance, examining the network for a similar hierarchical organization as it can be found in the primary visual cortex \cite{van1983hierarchical, felleman1991distributed}.

We found that, in general, the larger the ablated network portion, the stronger the effect on the classification performance. However, this effect greatly varies across different depths of the network. The results show that some layers are universally more important for the classification task than other layers. However, this effect varies across specific classes. We further investigated the possibilities to repair the inflicted damage by training the damaged network, in order to recover the original classification accuracy. Most of the negative effect of ablations on the classification accuracy could be recovered within a single episode of recovery training, even in cases of severe structural damage (up to 80\% of ablated filters within a single convolutional layer). 

Interestingly, for both networks, we found that ablations, despite having a general negative effect on the overall classification performance of the networks, consistently showed positive effects on the classification performance for specific classes. This raises the notion that the structure of a trained network may be purposefully manipulated to increase its classification performance beyond the local optimum that was reached during training.

\section{Related Work}
The basic idea of an ablation, i.e. removing trainable weights from a trained ANN, is also used when networks are pruned to reduce their size and computational cost. Pruning speeds up training and inference, while as much of the original performance as possible is retained. The idea is that some parameters of a trained network contribute very little or not at all to the output of the network and are therefore negligible and can be removed \cite{lecun1990optimal}. Recent research on pruning state-of-the-art CNNs, like the VGG-16 or the ResNet-110, focused on the optimization of a network's structure by removing kernels and entire filters \cite{molchanov2016pruning, li2016pruning} and methods to find an appropriate ranking of units to tackle the simple but challenging combinatorial optimization problem of how to choose the combination of units to be removed for the best results \cite{anwar2017structured, louizos2017learning, theis2018faster}. We aim to utilize the approach of ablations not merely to optimize the size and the speed of an ANN, but to gain insights about the structure and organization of the represented knowledge within the network, providing transparency and interpretability of the network's behavior. This objective is closely related to the question of how a network reaches its decisions and what the most important factors for this decision making process are. Some recent work on this matter demonstrated how the contribution of a network's input elements to its decision could be explained by means of Deep Taylor Decomposition \cite{montavon2017explaining} or Gradient-weighted Class Activation Mapping (Grad-CAM) \cite{selvaraju2017grad, chattopadhay2018grad}. Another recent example, which focused on the processes within a network rather than on the input, showed how latent representations within CNNs are stored in individual hidden units that align with a set of humanly interpretable semantic concepts \cite{bau2017network}. One of the most recent neuroscience-inspired contributions utilized ablations to demonstrate the relation between a network's capability to generalize a classification task and its reliance on class-selective single units within the network. Specifically, networks which generalize well contain less class-selective units than networks that merely memorize the dataset presented during training \cite{morcos2018importance}. 
\section{Methods}
In this study, we investigated two fundamentally different neural network architectures trained on two different datasets of varying complexity.
\subsection{Single and Pairwise Unit Ablations in a Shallow MLP}
In the first part of the study, we trained a small and shallow MLP to recognize handwritten digits using the MNIST dataset \cite{lecun-mnisthandwrittendigit-2010}. The input layer of the network comprises 784 units, which correspond to the 28x28 pixels of the input images. The network has two hidden layers with 20 and 10 hidden units respectively. ReLU activation was chosen for all hidden units. The output layer contains 10 units with softmax activation, which correspond to the 10 classes of the dataset. The network was trained for 100 epochs on 60,000 images of the training set and reached an accuracy of 94.64\% on the 10,000 images of the test set. After training, ablations of single units were performed by manually setting the weights of all incoming connections to zero, essentially preventing any kind of information flow through this unit. Since we trained the network without biases, zeroing a unit's incoming weights is equivalent to removing the unit from the network altogether. In order to investigate the effect of the ablation, we evaluated the performance of the network on the test set and compared the result with the original accuracy of the undamaged network. We used t-SNE \cite{maaten2008visualizing} on the complete 10,000 images of the test set to visualize the effects of the ablations.
\subsection{Ablations in the VGG-19} \label{ssec:vgg19_abl}
Second, we investigated the VGG-19 network with batch normalization, which was pre-trained on the ImageNet dataset \cite{imagenet_cvpr09}, as a representative of today's state-of-the-art CNNs for object recognition tasks. The VGG-19 has 19 layers with learnable weights, 16 convolutional and 3 fully connected layers. Because of its size, it allows for depth resolved investigations of the effects of ablations. Details about the dataset, the architecture of the network and the training process can be found in \cite{simonyan2014very}. The ImageNet dataset used for this study consists of 1,000 categories with a total of 1.2 million images in the training set and 50 images per category in the validation set. We performed ablations of groups of similar filters with increasing proportions (1\%, 5\%, 10\% and 25\%) relative to the total number of filters in each of the convolutional layers of the network. It is to be noted that due to the increasing sizes of the different convolutional layers, the same proportion may correspond to a different number of ablated filters. The similarity between filters within a group was calculated based on the absolute euclidean distance of the normalized filter weights. Similar to ablations of single units in the MLP, ablations were performed by manually setting the weights and biases of all incoming connections of a filter to 0, effectively eliminating any activation of that filter. The effect of ablations was evaluated by testing the classification performance of the network on the validation dataset using the top-1 and top-5 accuracy. 
\subsection{Recovery Training of the VGG-19} \label{ssec:vgg19_rec}
\begin{figure}[b!]
  \centering
  \vspace*{-10pt}
    \includegraphics[width=0.85\textwidth]{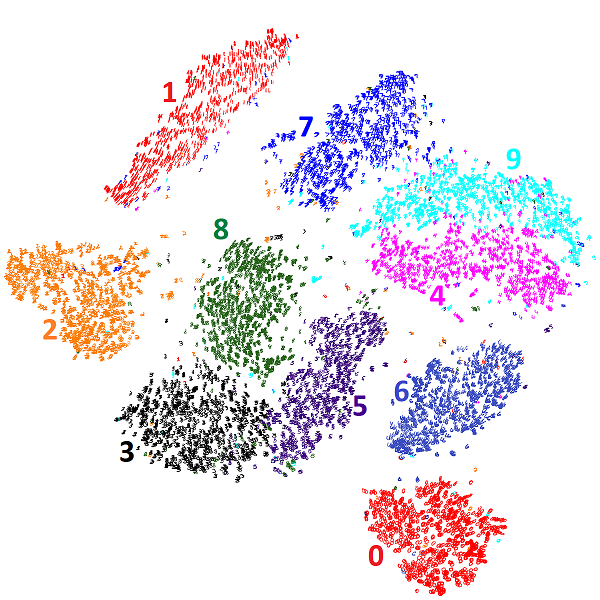}
    \caption{t-SNE visualization of the complete 10,000 digits of the MNIST test set.}
    \label{fig:tSNE_colored}
    \vspace*{-6pt}
\end{figure}
Following the observations made in the ablation study, we aimed to investigate the capability of the network to recover its original classification performance by subsequent recovery training of the damaged network. For this purpose, we performed ablations in the two most important layers for the classification task and retrained the damaged network. All the weights in the layers above the one in which the ablation was performed were frozen, forcing the network to adapt to the change of information flow through the deeper layers of the network. First, we investigated whether some filters are more important than others for recovering the classification performance. To this end, groups of filters with a proportion of 25\% of a layer's total number of filters were ablated in multiple instances of the network. After the ablation, the network was retrained with the training set for 5 epochs during which the top-5 accuracy was computed. Second, we investigated the impact of the amount of ablated filters within a layer on the network's recovery capabilities. For this purpose, we iteratively performed ablations of 25\% of the filters of one layer followed by recovery training, damaging the network further with each iteration. For each iteration, the filters to be ablated were chosen randomly and the recovery training was stopped after a minimum of 5 epochs when the top-5 accuracy did not improve by 0.05\% over the course of 2 epochs. Note that the choice of ablated filters was performed as a selection with replacement, i.e. two consecutive ablations of 25\% do not necessarily result in a total ablation of 50\%. This allows to perform more than 4 iterations of ablation and subsequent recovery training, slowly and gradually decreasing the amount of remaining filters in the damaged layer.
\section{Results}
\begin{figure}[b!]
    \centering
    \begin{subfigure}{0.51\textwidth}
        \includegraphics[width=\textwidth]{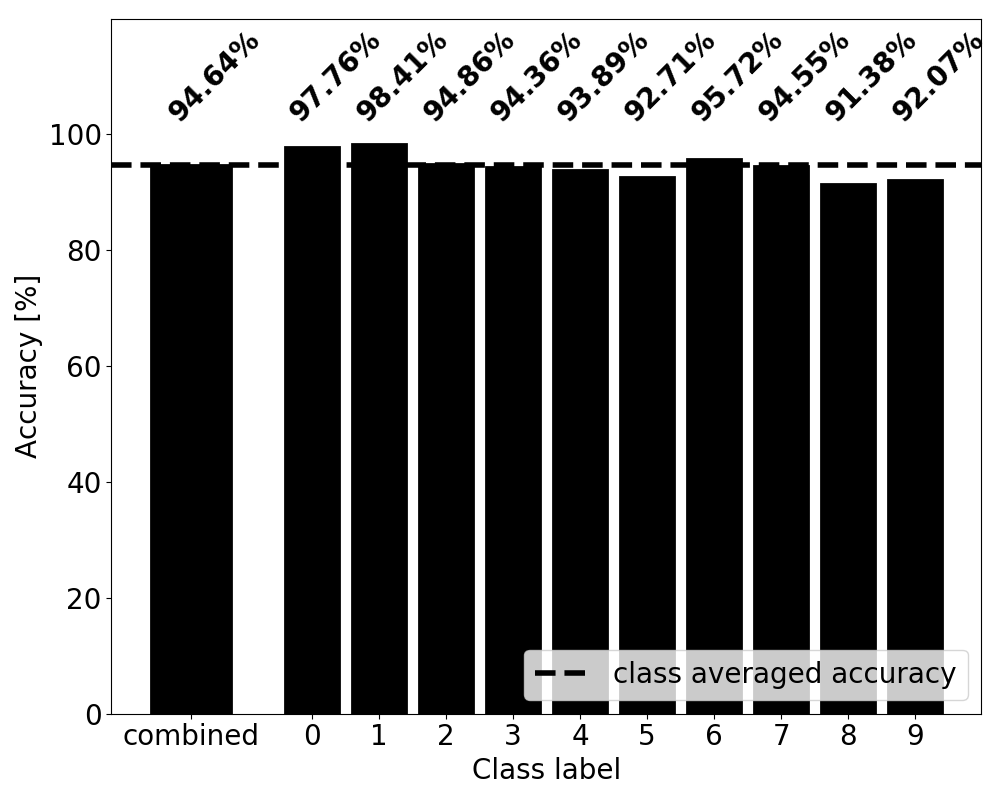}
    \end{subfigure}
    \hspace{0.01\textwidth}
    \begin{subfigure}{0.45\textwidth}
        \includegraphics[width=\textwidth]{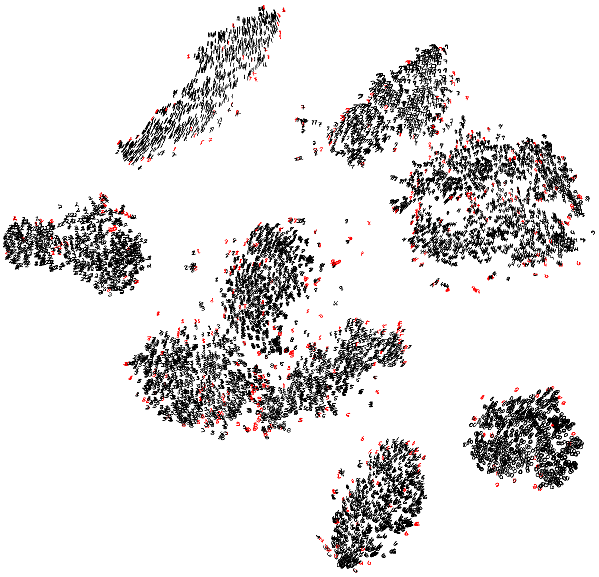}
    \end{subfigure}
    \caption{Overall accuracy, class-specific accuracy and t-SNE visualization of the trained MLP.}
    \label{fig:acc-tSNE_trained}
    \vspace*{-10pt}
\end{figure}

Figure \ref{fig:tSNE_colored} shows a t-SNE visualization of the 10,000 digits in the test set and serves as a basis for the visual evaluation of the effects of ablations. As t-SNE tries to preserve the global and local structure of the data when embedding the original 784-dimensional dataset into the 2-dimensional space, it allows us to investigate whether this structure is represented in an organized manner in the network. The overall accuracy of the trained MLP on the test set was 94.6\% with a slight variation across the classes ranging from 91.4\% for class 8 to 98.4\% for class 1. Figure \ref{fig:acc-tSNE_trained} shows the overall classification accuracy of the MLP, its class-specific variation and the corresponding t-SNE plot. The black and red digits correspond to the correctly and incorrectly classified input images. 

\subsection{Single Unit Ablations in a Shallow MLP}
We found that the ablations of single units affected the accuracy in different ways. In general, the overall accuracy decreased, whereas the effect on single classes differed for specific ablations. 

\begin{figure}[t]
    \centering
    \begin{subfigure}{0.51\textwidth}
        \includegraphics[width=\textwidth]{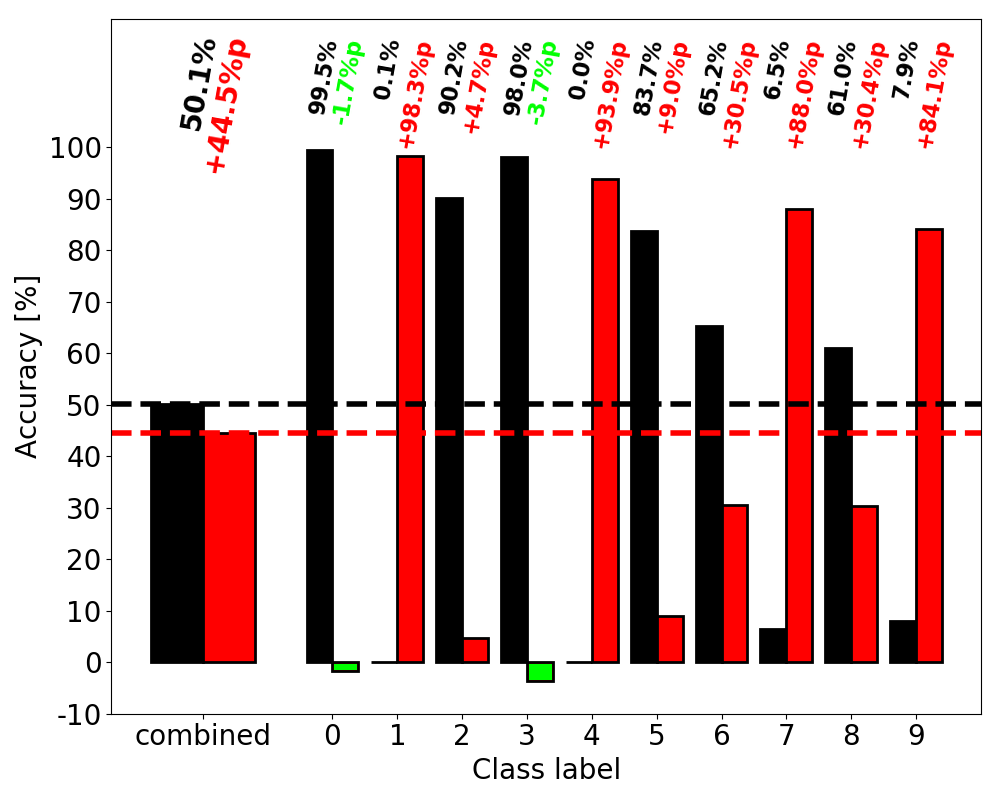}
    \end{subfigure}
    \hspace{0.01\textwidth}
    \begin{subfigure}{0.45\textwidth}
        \includegraphics[width=\textwidth]{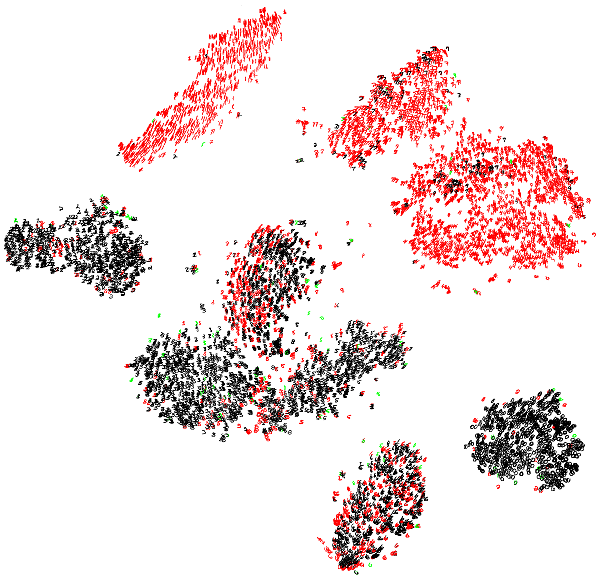}
    \end{subfigure}
    \caption{Overall accuracy, class-specific accuracy and t-SNE visualization of the damaged MLP after the ablation of unit 12 in the first hidden layer. This unit is an example for the representation of features corresponding to many different classes}
    \label{fig:acc-tSNE_ko12}
    \vspace*{-10pt}
\end{figure}

Figure \ref{fig:acc-tSNE_ko12} shows the effects of the ablation of unit 12 in the first hidden layer of the MLP, which resulted in the highest drop of overall accuracy of 44.5\%p for a single ablated unit.
The heights of the black and red/green bars correspond to the amount of correctly and incorrectly classified digits after the ablation, respectively. However, the red colored digits do not contain the digits that were incorrectly classified by the undamaged network and only display the change of the classification performance as a result of the ablation. Green bars with a negative value correspond to the amount of correctly classified digits after the ablation, which were incorrectly classified by the undamaged network, thus representing an improvement of the classification performance. The network lost its ability to correctly classify most digits of the classes 1, 4, 7 and 9 with a drop in class-specific accuracy of more than 80\%p. The effects on the classes 6 and 8 are less severe with a drop in class-specific accuracy of around 30\%p, while the effect on all other classes is smaller than 10\%p. The t-SNE plot suggests that this unit represents certain features in the data that are shared across classes, as the majority of incorrectly classified digits are located close to each other in the upper part of the plot. Figure \ref{fig:acc-tSNE_ko16} shows a similar representation for an ablation of unit 16, where most of the incorrectly classified units are found in the bottom right part of the t-SNE plot.

\begin{figure}[t!]
    \centering
    \begin{subfigure}{0.51\textwidth}
        \includegraphics[width=\textwidth]{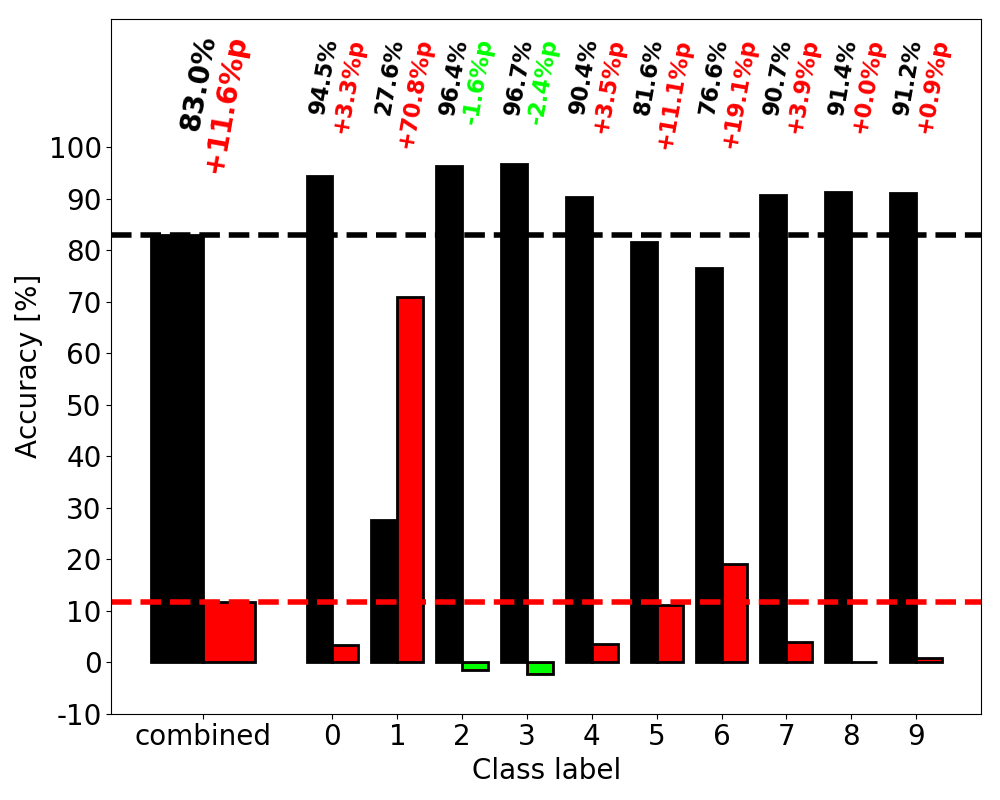}
    \end{subfigure}
    \hspace{0.01\textwidth}
    \begin{subfigure}{0.45\textwidth}
        \includegraphics[width=\textwidth]{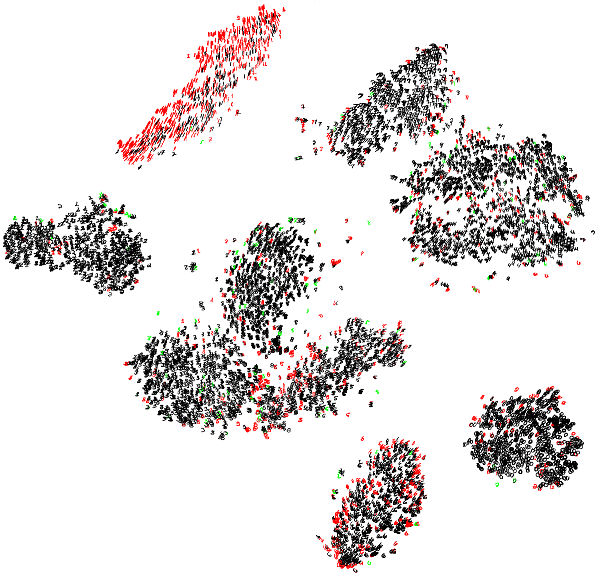}
    \end{subfigure}
    \caption{Overall accuracy, class-specific accuracy and t-SNE visualization of the damaged MLP after the ablation of unit 19 in the first hidden layer. This unit is an example for the selective representation features distinct to a single class.}
    \label{fig:acc-tSNE_ko19}
     \vspace*{-10pt}
\end{figure}

Figure \ref{fig:acc-tSNE_ko19} shows the effects of the ablation of unit 19 in the first hidden layer of the MLP, which resulted in a drop of overall accuracy of 11.6\%p. In contrast to unit 12, this unit seems to represent features distinct to a single class, as the effect on the class-specific accuracy for class one is much stronger than for all other classes. Although this unit is easy to interpret as it seems to represent features of a single class almost exclusively, it is not more important for the classification task than other units, in terms of how strongly its ablation affects the overall classification performance. This result is consistent with previous investigations on the interpretability and importance of single units of an MLP classifier \cite{morcos2018importance}.

\begin{figure}[!t]
    \centering
    \begin{subfigure}{0.51\textwidth}
        \includegraphics[width=\textwidth]{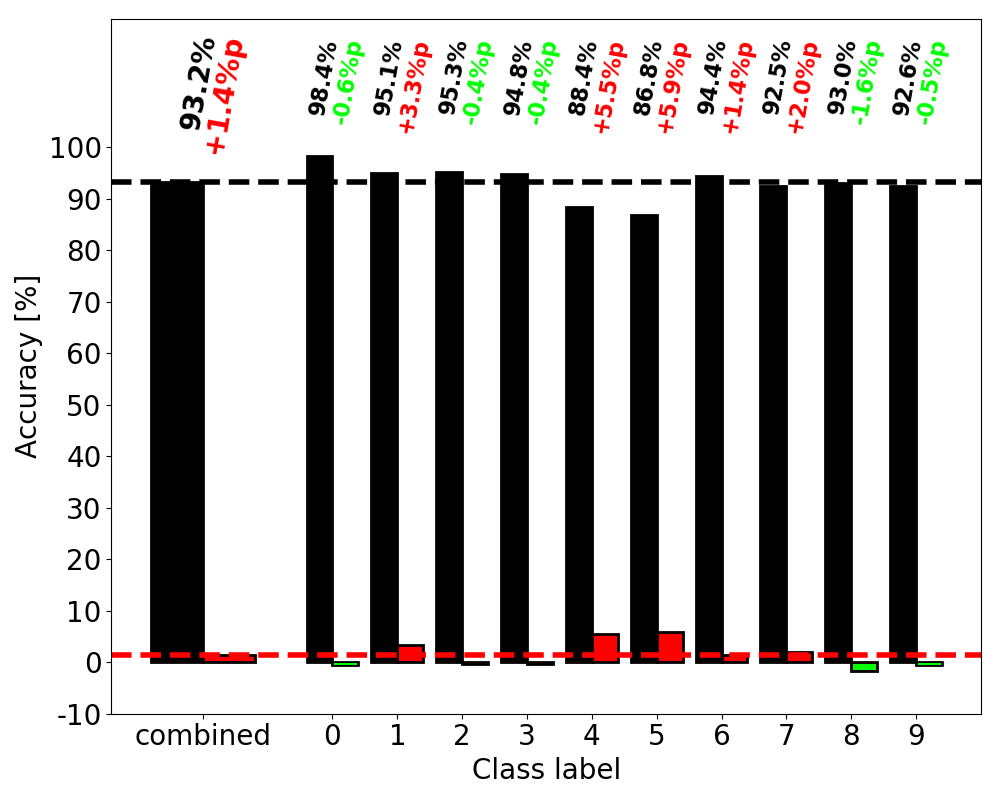}
        \label{fig:acc}
    \end{subfigure}
    \hspace{0.01\textwidth}
    \begin{subfigure}{0.45\textwidth}
        \includegraphics[width=\textwidth]{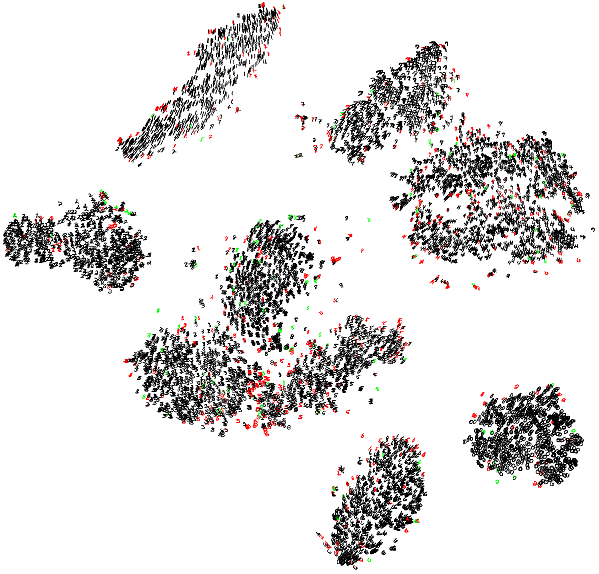}
        \label{fig:tSNE}
    \end{subfigure}
    \caption{Overall accuracy, class-specific accuracy and t-SNE visualization of the damaged MLP after the ablation of unit 6 in the first hidden layer. This unit is an example for a negligible contribution to the classification task and could be pruned to optimize network size.}
    \label{fig:acc-tSNE_ko6}
    \vspace*{-10pt}
\end{figure}

Figure \ref{fig:acc-tSNE_ko6} shows the effects of the ablation of unit 6 in the first hidden layer of the MLP, which resulted in a drop of overall accuracy of only 1.4\%p. This unit seems to play only a minor role in the classification task as the effect of its ablation on the networks accuracy is small. We found that 4 out of the 20 units in the first hidden layer, unit 6, 11, 13 and 18, showed similar effects which makes them top candidates for pruning, if one would want to optimize the size of the network (c.f. Figure \ref{fig:acc-tSNE_ko6111318}).

\begin{figure}[t!]
    \centering
    \begin{subfigure}{0.51\textwidth}
        \includegraphics[width=\textwidth]{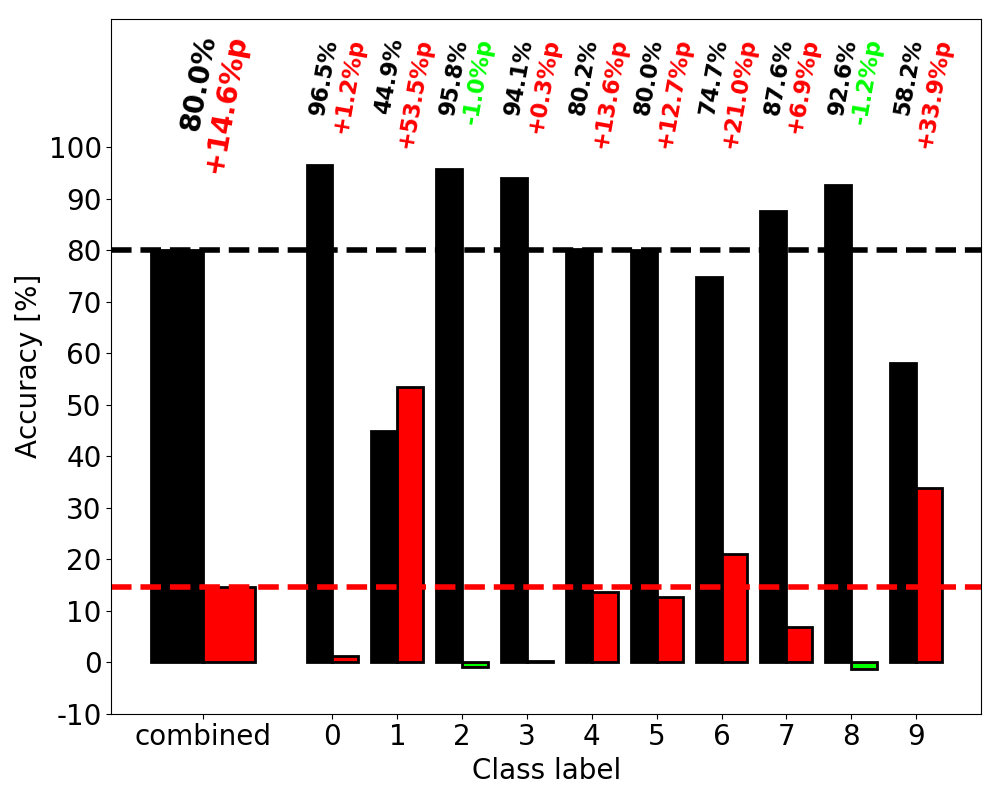}
    \end{subfigure}
    \hspace{0.01\textwidth}
    \begin{subfigure}{0.45\textwidth}
        \includegraphics[width=\textwidth]{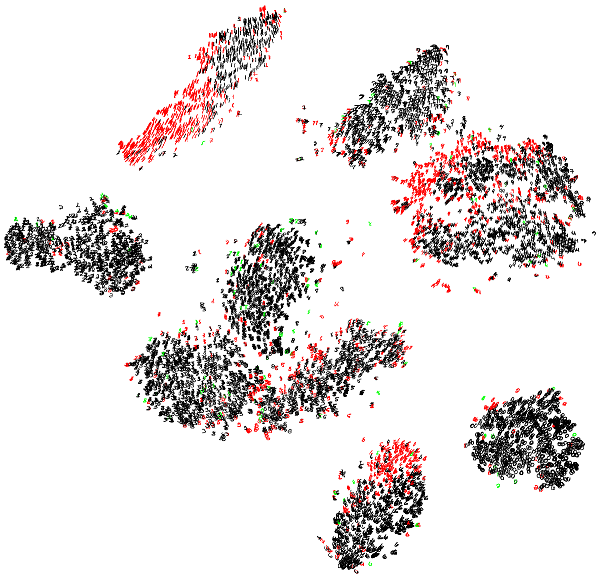}
    \end{subfigure}
    \hspace{0.01\textwidth}
    \begin{subfigure}{0.48\textwidth}
        \includegraphics[width=\textwidth]{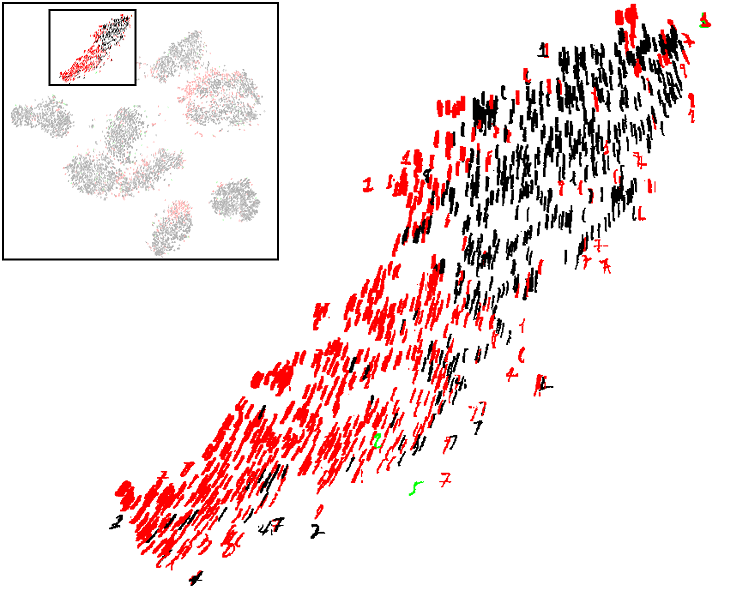}
    \end{subfigure}
    \hspace{0.01\textwidth}
    \begin{subfigure}{0.48\textwidth}
        \includegraphics[width=\textwidth]{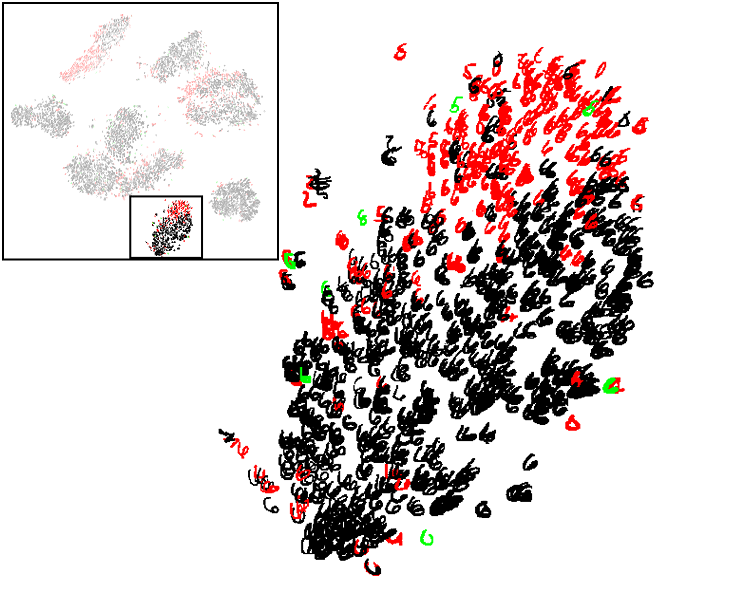}
    \end{subfigure}
    \caption{Overall accuracy, class-specific accuracy and t-SNE visualization of the damaged MLP after the ablation of unit 20 in the first hidden layer. This unit is an example for the representation of features that are distinct to a subset of digits within different classes.}
    \label{fig:acc-tSNE_ko20}
    \vspace*{-10pt}
\end{figure}

Figure \ref{fig:acc-tSNE_ko20} shows the effects of the ablation of unit 20 in the first hidden layer of the MLP, which resulted in a drop of overall accuracy of 14.6\%p. This unit seems to represent features corresponding to subtle and smoothly changing characteristics distinct to the classes 1, 6 and 9. The t-SNE visualization reveals that most of the incorrectly classified digits within a class can be found close to each other rather than evenly distributed across the whole class.
\begin{figure}[t!]
    \centering
    \begin{subfigure}{0.49\textwidth}
        \includegraphics[width=\textwidth]{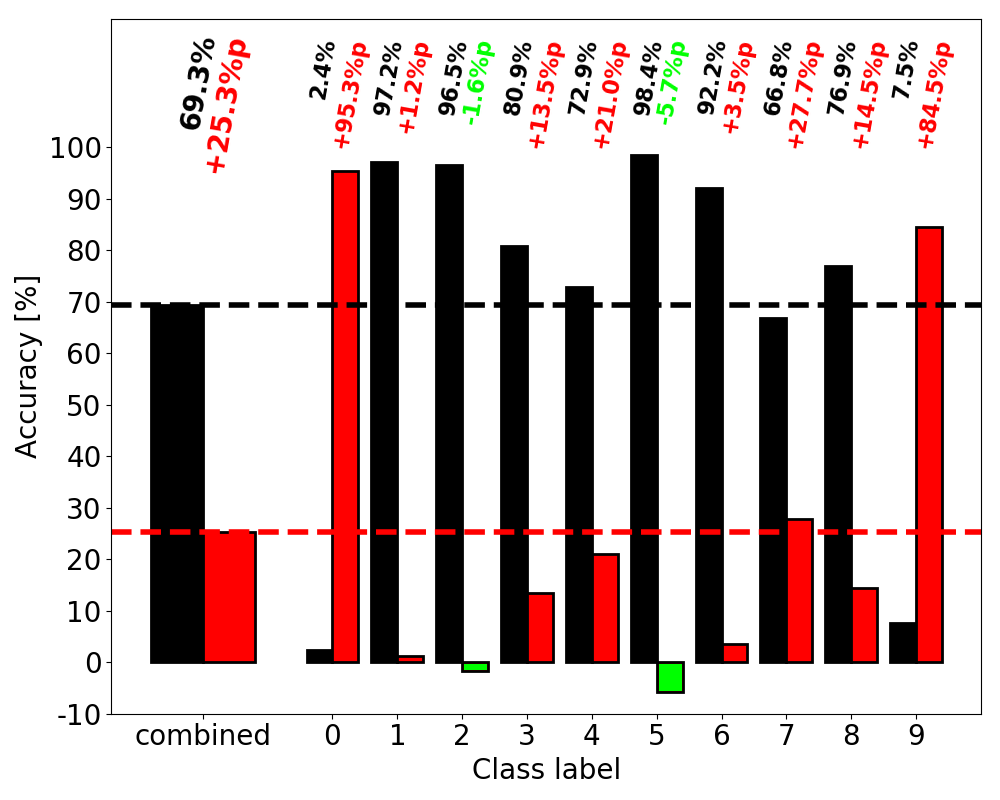}
    \end{subfigure}
    \hspace{0.01\textwidth}
    \begin{subfigure}{0.47\textwidth}
        \includegraphics[width=\textwidth]{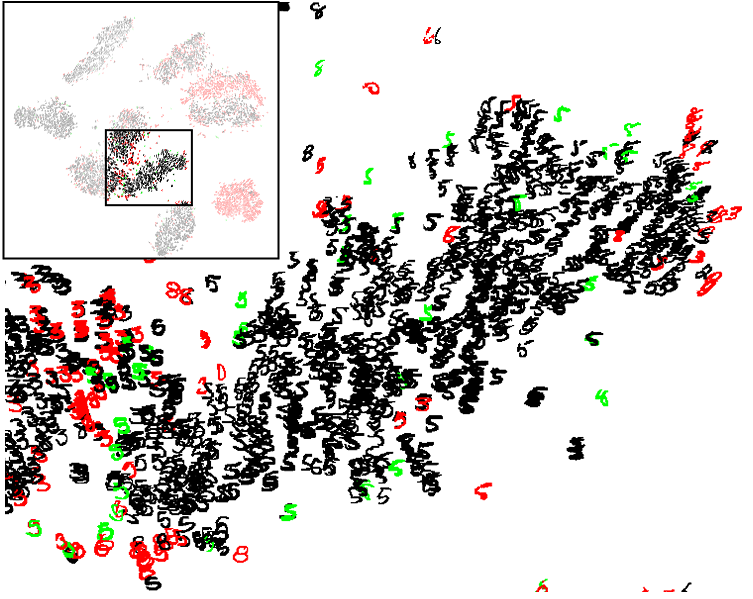}
    \end{subfigure}
    \caption{Overall accuracy, class-specific accuracy and t-SNE visualization of the damaged MLP after the ablation of unit 3 in the first hidden layer. This unit shows the strongest positive effect of an ablation, i.e. the increase of the class-specific accuracy of class 5.}
    \label{fig:acc-tSNE_ko3}
    \vspace*{-10pt}
\end{figure}

Figure \ref{fig:acc-tSNE_ko3} shows the effects of the ablation of unit 3 in the first hidden layer of the MLP, which resulted in a drop of overall accuracy of 25.4\%p but showed an increase of the class-specific accuracy of 5.7\% for class 5, which is the strongest positive effect of all units in the first hidden layer. This observation is consistent across ablated single units, i.e. the damaged network would correctly classify some digits that were incorrectly classified by the undamaged network. These observations hint to a trade-off made while fitting the weights of the network via backpropagation, in which the recognition of a small amount of digits is sacrificed for a much larger amount of other digits. However, this raises the question whether the classification performance of a network can be increased beyond its trained capabilities by selectively ablating single connections to achieve the desired increase in accuracy without suffering from the negative effects.

Following the observations of the ablations, we aimed to find characteristics of the single units which correlate with the drop in the overall accuracy after ablation of these units. Such characteristics would allow to describe the importance of single units for the classification task without the necessity to perform a functional test. We found that the degree to which the distribution of the incoming weights of a particular unit after training differs from the randomly initialized normal distribution of weights before training is a good indication of the unit's importance for the classification task. We quantified this difference by the p-value of the Mann-Whitney U test, a non-parametric statistical test, which determines whether two independent observations were sampled from the same distribution. The p-value indicates the likelihood of both distributions to be the same ($p = 1$) or to be different from each other ($p \rightarrow 0$). 
\begin{figure}[t!]
    \centering
    \begin{subfigure}{0.48\textwidth}
        \includegraphics[width=\textwidth]{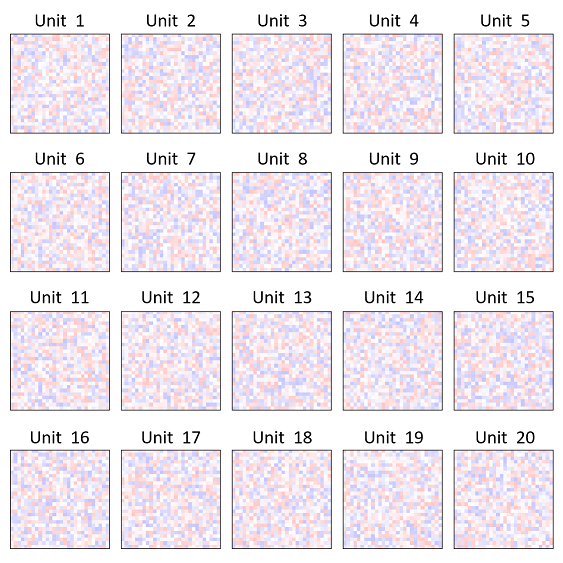}
    \end{subfigure}
    \hspace{0.01\textwidth}
    \begin{subfigure}{0.48\textwidth}
        \includegraphics[width=\textwidth]{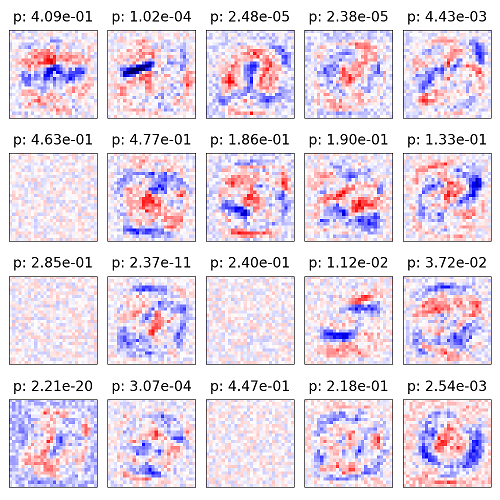}
    \end{subfigure}
    \caption{Comparison of the distributions of the incoming weights for the 20 single units in the first hidden layer before training (left) and after training (right).}
    \label{fig:weights_distrib}
    \vspace*{-10pt}
\end{figure}
Figure \ref{fig:weights_distrib} shows a comparison of the weight distributions of the single units in the first hidden layer before and after training. Each distribution is visualized as a normalized 28x28 matrix, the same dimensions as the input images, with red and blue entries indicating high positive and negative weight values, respectively. The p-values indicate the difference of the distributions on the right hand side compared to the left hand side. Note that the distributions of unit 6, 11, 13, and 18 did not change significantly during training (c.f. Figure \ref{fig:acc-tSNE_ko6111318}). 

\begin{figure}[t]
    \centering
    \begin{subfigure}{0.48\textwidth}
        \includegraphics[width=\textwidth]{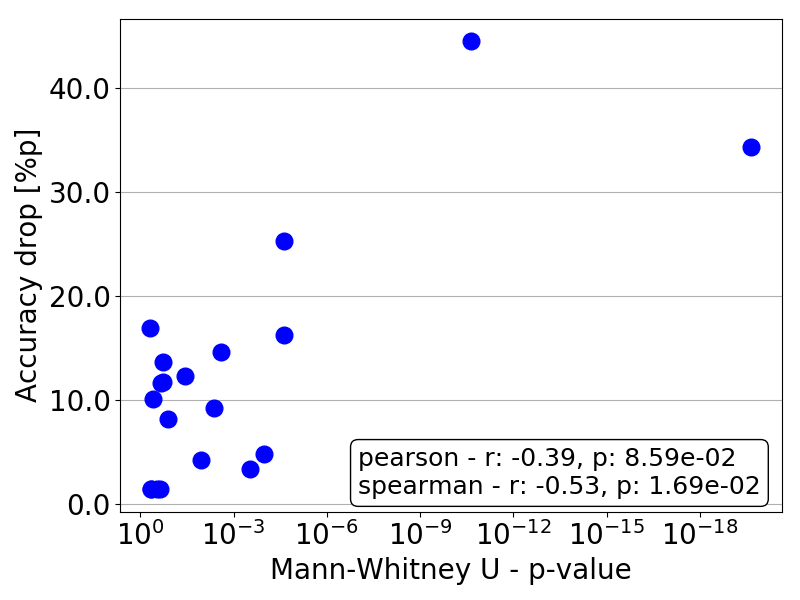}
    \end{subfigure}
    \hspace{0.01\textwidth}
    \begin{subfigure}{0.48\textwidth}
        \includegraphics[width=\textwidth]{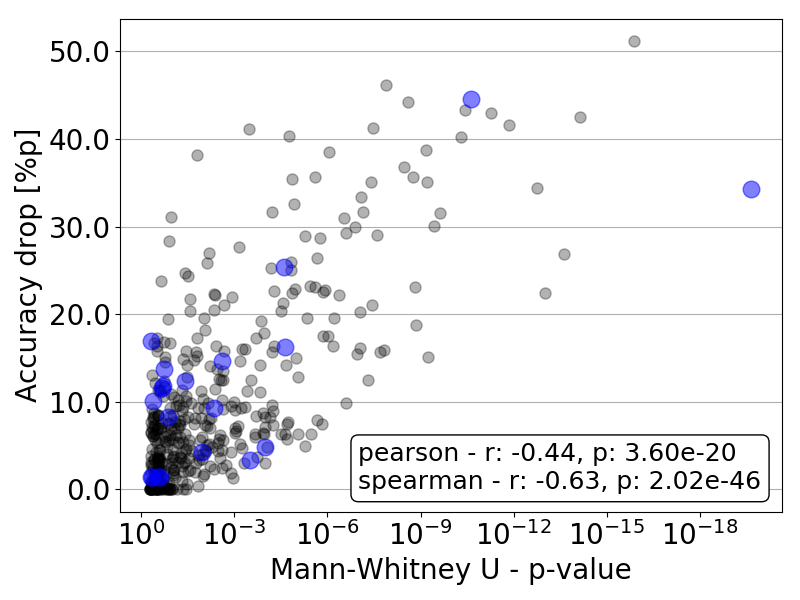}
    \end{subfigure}
    \caption{Correlation of the Mann-Whitney U's p-value with the drop in accuracy after ablation of a single unit in the first hidden layer.}
    \label{fig:acc_corr}
    \vspace*{-10pt}
\end{figure}

Figure \ref{fig:acc_corr} shows the Pearson and Spearman correlation of the Mann-Whitney U's p-value and the drop in accuracy after ablation. The left side shows 20 samples corresponding to the 20 units in the first hidden layer of the network from which the previous results were generated. In order to verify that the observed correlation is not a result of the random initialization of the network, we trained 20 more networks with different initializations and calculated the correlation coefficients for all 400 units within the first hidden layers of the 20 networks (c.f. Figure \ref{fig:acc_corr}, right side). The results suggest that, in general, the more a single unit's distribution of incoming weights changes during training, the more important this unit is for the overall classification performance.
\begin{figure}[t!]
    \centering
    \begin{minipage}[b]{0.47\textwidth}
        \includegraphics[width=\textwidth]{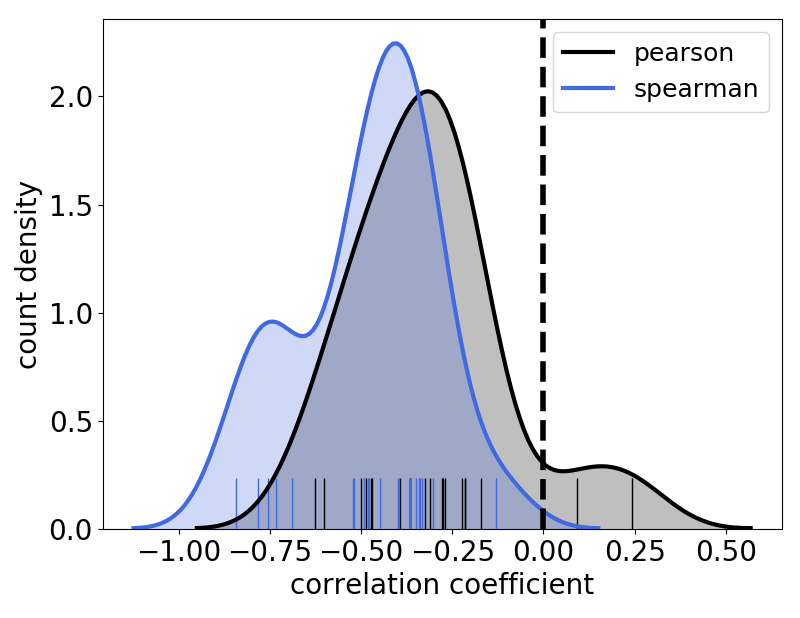}
        \caption{Distributions of the calculated Pearson and Spearman correlation coefficients for the 20 networks.}
        \label{fig:corr_distrib}
    \end{minipage}
    \hspace{0.01\textwidth}
    \begin{minipage}[b]{0.49\textwidth}
        \includegraphics[width=\textwidth]{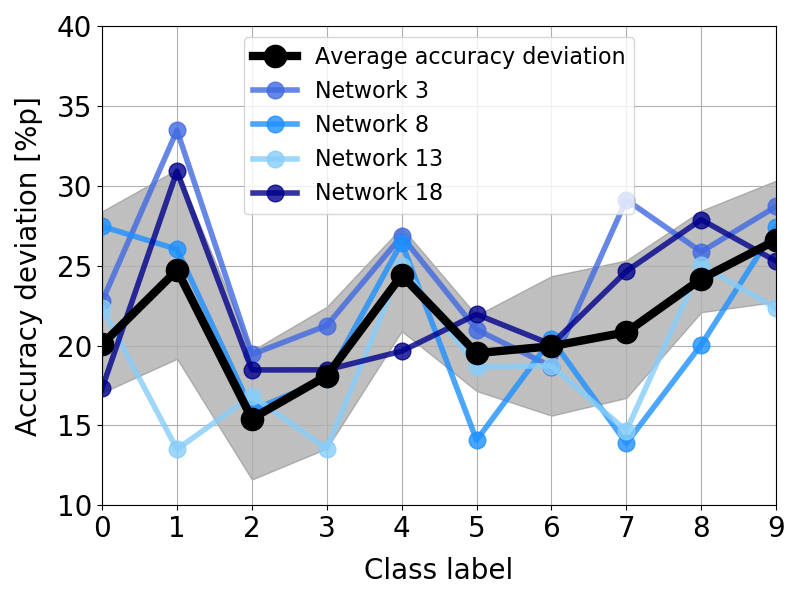}
        \caption{Class-specific averaged deviation across the 20 networks of the accuracy drop after ablations.}
        \label{fig:class_spec_pop_avr}
    \end{minipage}
    \vspace*{-10pt}
\end{figure}

Figure \ref{fig:corr_distrib} shows a kernel density estimated distribution of the calculated Pearson and Spearman correlations from all 20 networks and, except for two Pearson coefficients, supports the average trend shown in Figure \ref{fig:acc_corr}. This observation may prove useful for pruning neural networks. Units may be pruned based on the distributions of their incoming weights, thus, reducing the computational cost of repeatedly testing a pruned network on a large dataset.

We wondered whether the representation of some classes within the networks is more selective than for other classes. For this purpose, we tested if the drop of the class-specific accuracy after an ablation is similar for all units within a network or if it shows a strong deviation. A high deviation would mean that some units within the network strongly represent a particular class while other units do not. This would suggest that this class is represented somewhat localized in the network rather than evenly distributed across all units. Therefore, for each of the 20 networks, we computed the class-specific drop in accuracy for all 20 single unit ablations in the first hidden layer and calculated the standard deviation. We further calculated the mean of this class-specific accuracy deviation averaged across all 20 networks in order to compare the deviations of the single networks to the population mean. Figure \ref{fig:class_spec_pop_avr} shows the population averaged accuracy deviation and four examples of a single network accuracy deviation. The black line corresponding to the population averaged accuracy deviation shows that some classes are represented more selectively than other classes. For instance, the classes 1 and 4 have a much higher deviation than class 2, suggesting that, in general, class 2 is much more evenly represented across the first hidden layer than the classes 1 and 4. However, this trend is not universal for all 20 networks indicated by the accuracy deviations of the networks. The fact that the blue lines cross the population average suggests that, despite the general trend, the selectivity of the representation of the 10 classes is rather unique to each network. This means that some networks develop a more selective representation for some classes than others.
\subsection{Pairwise Unit Ablations in a Shallow MLP}
In addition to single unit ablations, we performed pairwise unit ablations in the first hidden layer of the MLP to investigate the feature representations for redundancies, i.e. whether the effects of pairwise unit ablations are stronger than the sum of the corresponding single unit ablations. In this case, the network retains its capability to correctly classify some specific classes after a single unit ablation as another unit still represents the corresponding features sufficiently well. However, the pairwise unit ablation of both of these units causes the network to incorrectly classify those classes that were correctly classified in the case of the single unit ablations, as there are no more units left that redundantly represent the necessary features. 
\begin{figure}[b!]
    \centering
    \begin{subfigure}{0.51\textwidth}
        \includegraphics[width=\textwidth]{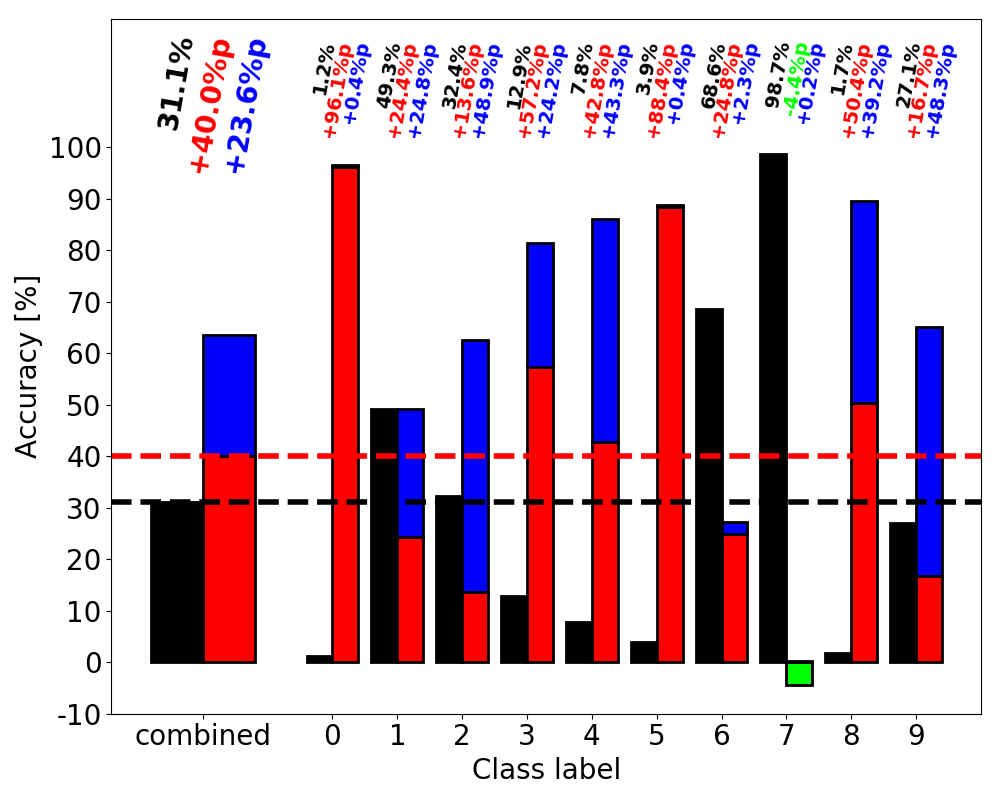}
    \end{subfigure}
    \hspace{0.01\textwidth}
    \begin{subfigure}{0.45\textwidth}
        \includegraphics[width=\textwidth]{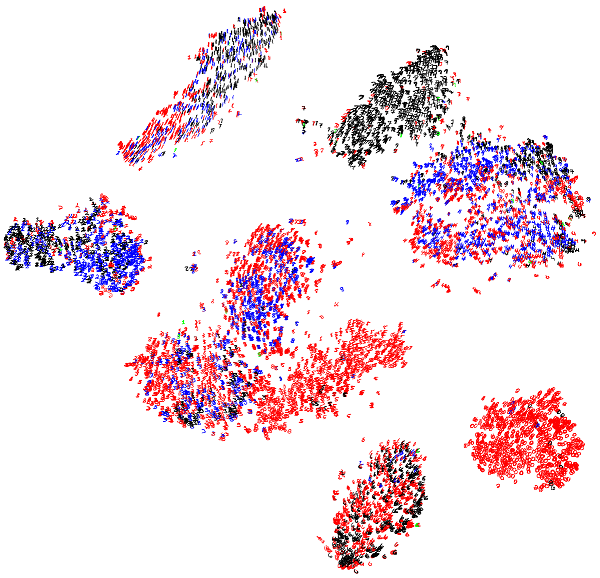}
    \end{subfigure}
    \caption{Overall accuracy, class-specific accuracy and t-SNE visualization of the damaged MLP after the ablation of units 4 and 16 in the first hidden layer. The pairwise ablation of these units had the strongest effect exceeding the summed effects of the corresponding single unit ablations.}
    \label{fig:acc-tSNE_ko416}
    \vspace*{-10pt}
\end{figure}
\begin{figure}[t!]
    \centering
    \begin{subfigure}{0.51\textwidth}
        \includegraphics[width=\textwidth]{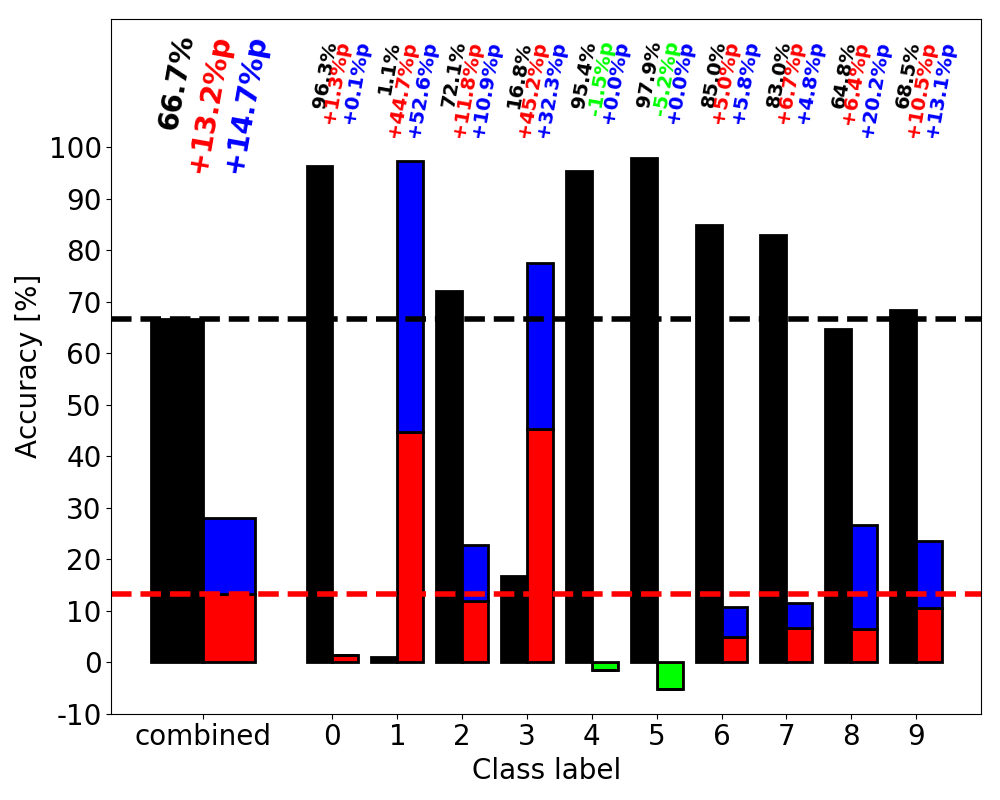}
    \end{subfigure}
    \hspace{0.01\textwidth}
    \begin{subfigure}{0.45\textwidth}
        \includegraphics[width=\textwidth]{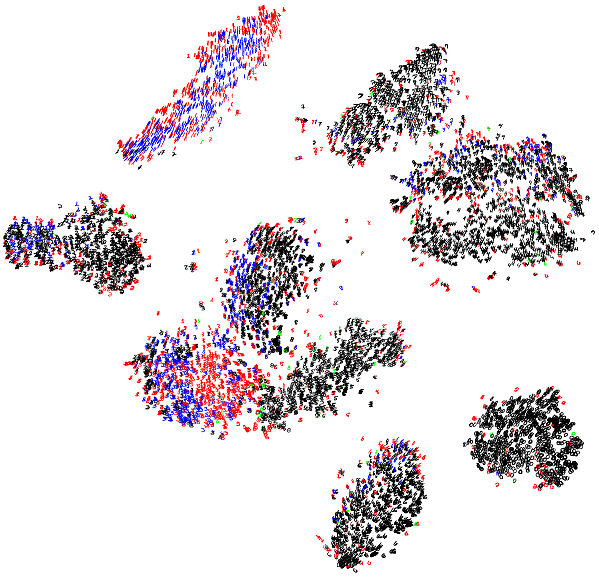}
    \end{subfigure}
    \hspace{0.01\textwidth}
    \begin{subfigure}{0.48\textwidth}
        \includegraphics[width=\textwidth]{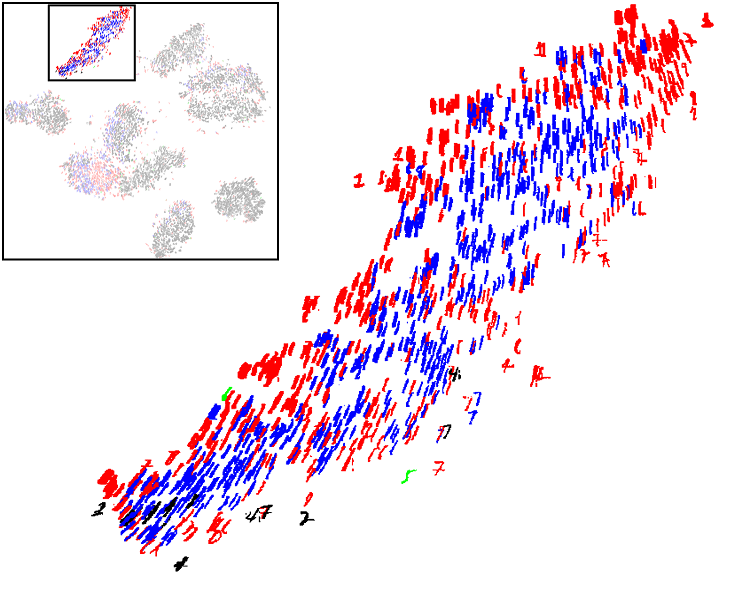}
    \end{subfigure}
    \hspace{0.01\textwidth}
    \begin{subfigure}{0.48\textwidth}
        \includegraphics[width=\textwidth]{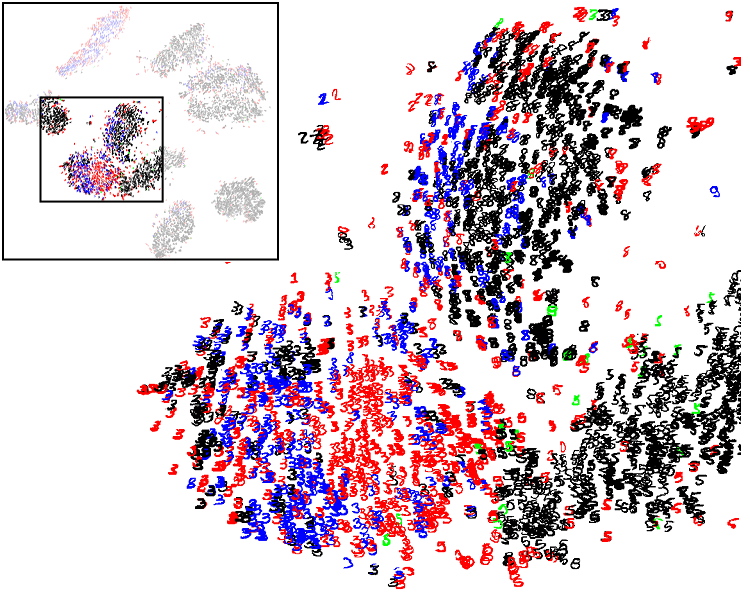}
    \end{subfigure}
    \caption{Overall accuracy, class-specific accuracy and t-SNE visualization of the damaged MLP after the pairwise unit ablation of units 5 and 10 in the first hidden layer. Note that the positive effect on class five is stronger after the pairwise unit ablation than the summed effects after the corresponding single unit ablations (c.f. Figure \ref{fig:acc_ko510})}
    \label{fig:acc-tSNE_ko510}
    \vspace*{-14pt}
\end{figure} 
Figure \ref{fig:acc-tSNE_ko416} shows the effects of the pairwise unit ablation of units 4 and 16 in the first hidden layer of the MLP, causing the strongest observed effect to exceed the sum of the corresponding single unit ablations. The height of the black, red/green and blue bars correspond to the amount of objects correctly classified after the pairwise unit ablation, the amount of objects incorrectly classified after either corresponding single unit ablation and the amount of objects incorrectly classified only after the pairwise unit ablation. The digits in the t-SNE plot are colored accordingly. As a direct comparison to the single unit ablations of unit 12 (c.f. Figure \ref{fig:acc-tSNE_ko12}) and 19 (c.f. Figure \ref{fig:acc-tSNE_ko19}), Figure \ref{fig:acc-tSNE_ko1219} shows the pairwise unit ablation of units 12 and 19. The pairwise unit ablation has a strong effect on class 6, for which more than 50\% of the digits are incorrectly classified as a result of the pairwise unit ablation but were correctly classified after either corresponding single unit ablation. The t-SNE visualization shows that the digits corresponding to that redundant representation are more or less evenly distributed across the class. Note that the positive effect on class 3 is stronger than on either single unit ablation, however this effect does not exceed the summed effects of both single unit ablations. Contrary to the units 12 and 19, the t-SNE visualization for the pairwise unit ablation of the units 5 and 10 suggests that the redundantly represented features correspond to the local structure of the data (c.f. Figure \ref{fig:acc-tSNE_ko510}). The blue colored digits within class 1 and 3 are clustered together, rather than being evenly distributed across the whole class. Interestingly, we found that the positive effect of the ablation can exceed the effects of the corresponding single unit ablations. Even though the pairwise unit ablation shows strong class-specific negative effects, the positive effect on class 5, improving the amount of correctly classified objects by 5.16\%p, exceeds the summed effects of the corresponding single unit ablations of 3.14\%p for unit 5 and 0.45\%p for unit 10 (c.f. Figure \ref{fig:acc_ko510}).
\subsection{Ablations in the VGG-19}
\begin{figure}[t!]
    \centering
    \begin{subfigure}{0.48\textwidth}
        \includegraphics[width=\textwidth]{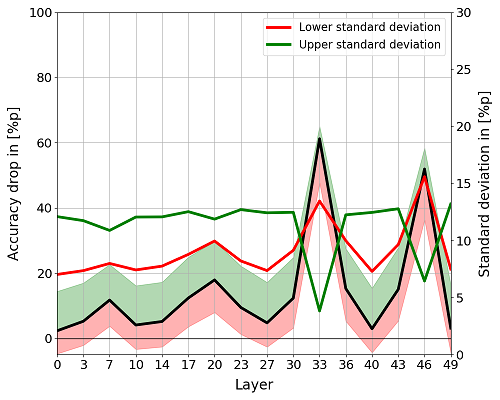}
    \end{subfigure}
    \hspace{0.01\textwidth}
    \begin{subfigure}{0.48\textwidth}
        \includegraphics[width=\textwidth]{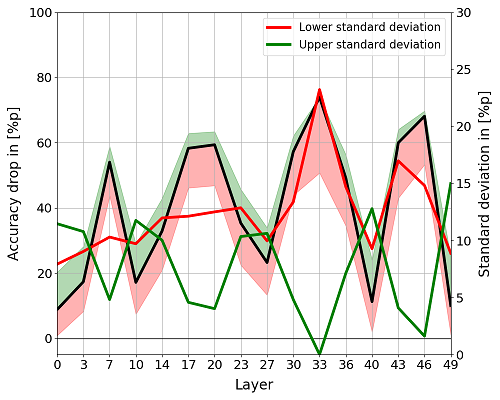}
    \end{subfigure}
    \caption{Effect on the top-1 accuracy of ablations of different amounts (left: 10\% of layer filters, right: 25\% of layer filters) in all convolutional layers.}
    \label{fig:vgg_acc_top1}
    \vspace*{-10pt}
\end{figure}

\begin{figure}[t]
    \centering
    \begin{subfigure}{0.48\textwidth}
        \includegraphics[width=\textwidth]{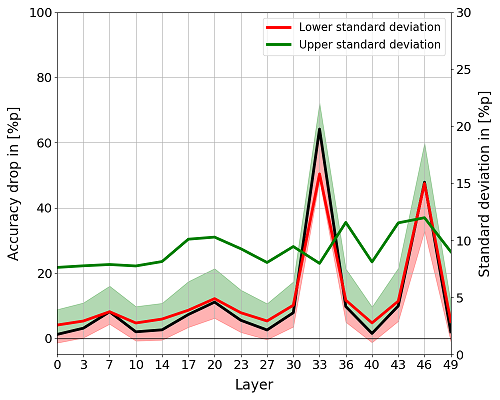}
    \end{subfigure}
    \hspace{0.01\textwidth}
    \begin{subfigure}{0.48\textwidth}
        \includegraphics[width=\textwidth]{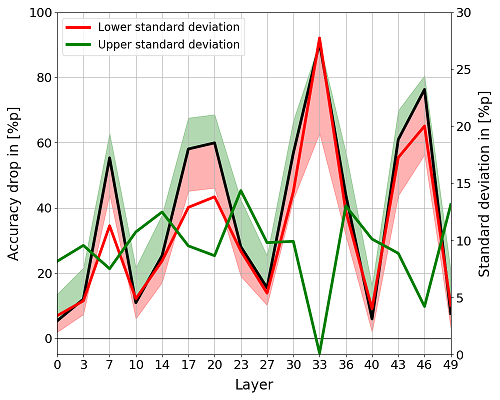}
    \end{subfigure}
    \caption{Effect on the top-5 accuracy of ablations of different amounts (left: 10\% of layer filters, right: 25\% of layer filters) in all convolutional layers.}
    \label{fig:vgg_acc_top5}
    \vspace*{-10pt}
\end{figure}

Complementary to the investigation of single units in a shallow MLP, we investigated the VGG-19 as a representative of state-of-the-art CNNs for image classification tasks. The VGG-19 is much larger and deeper than the previously investigated MLP and allows for depth resolved investigations of the effects of ablations. Similar to the importance of single units in the MLP, we found that some layers are more important for the classification task than other layers. Figure \ref{fig:vgg_acc_top1} and Figure \ref{fig:vgg_acc_top5} show the drop in top-1 and top-5 accuracy, respectively, for the ablation of 10\% (left side) and 25\% (right side) in all convolutional layers of the VGG-19.
The black curve shows the accuracy drop in each layer, averaged over all ablations performed in this layer. The number of ablations is equal to the number of filters in each particular layer, since each filter was chosen once as a reference for the choice of the 10\% and 25\% ablations based on filter similarity (c.f. section \ref{ssec:vgg19_abl}). The red and green curves and shaded areas correspond to the lower and upper standard deviation from the average accuracy drop. Layer 33 and 46 showed a significantly higher drop in the top-1 and top-5 accuracy compared to other layers. This effect is more distinct for the smaller amount of ablated filters (10\%) and becomes less pronounced for the larger amount (25\%). Concurrently, the effect of a larger amount of ablated filters has a stronger impact on some layers than on others. For instance, layers 7, 17, and 20 show a significantly stronger drop in both, top-1 and top-5 accuracy for 25\% of ablated filters compared to 10\% of ablated filters, while layer 40 is almost not affected at all. This observation is somewhat surprising as we expect the upper layers to be the most important layers, as they are supposed to represent more general features common to many classes, whereas lower layers are supposed to represent more class-specific features \cite{zeiler2014visualizing}. Additionally, the fact that some layers, e.g. layer 40, are largely unaffected by the increase of the proportion of ablated filters from 10\% to 25\% suggests that features represented in this layer may be redundantly represented in other layers or in other filters in the same layer, rendering ablations mostly harmless for the overall performance. Consistent with the observations of positive effects of ablations in the MLP study, the ablation of some filters in some layers of the VGG-19 showed an increase in top-1 accuracy indicated by the crossing of the zero-line of the red shaded area in Figure \ref{fig:vgg_acc_top1}.

\begin{figure}[!t]
    \centering
    \begin{subfigure}{0.48\textwidth}
        \includegraphics[width=\textwidth]{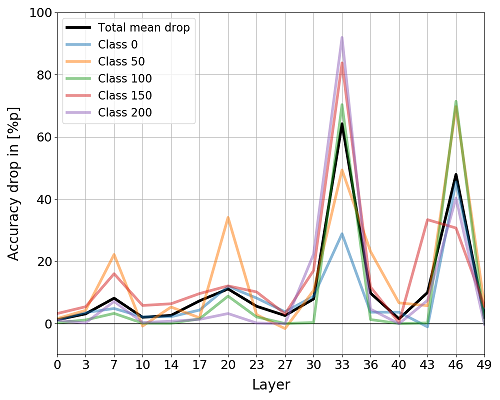}
    \end{subfigure}
    \hspace{0.01\textwidth}
    \begin{subfigure}{0.48\textwidth}
        \includegraphics[width=\textwidth]{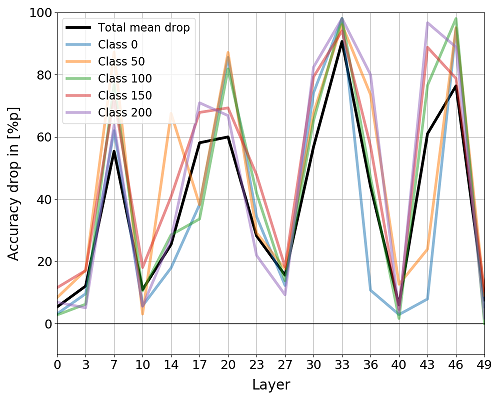}
    \end{subfigure}
    \caption{Examples for the variation of the class-specific effect of ablations on the top-5 accuracy for different amounts of ablated filters (left: 10\%, right: 25\%) in all convolutional layers.}
    \label{fig:vgg_acc_top5_class}
    \vspace*{-10pt}
\end{figure}

\begin{figure}[!t]
    \centering
    \begin{subfigure}{0.48\textwidth}
        \includegraphics[width=\textwidth]{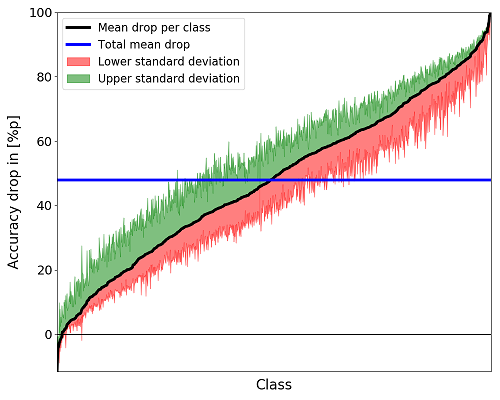}
    \end{subfigure}
    \hspace{0.01\textwidth}
    \begin{subfigure}{0.48\textwidth}
        \includegraphics[width=\textwidth]{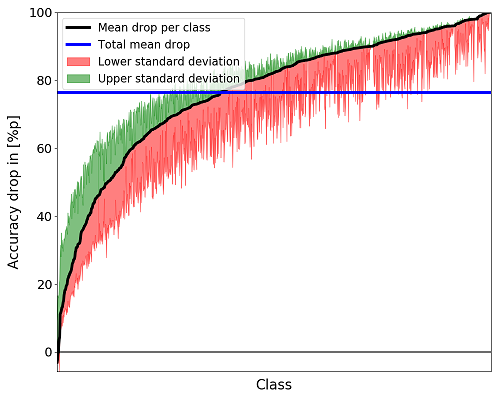}
    \end{subfigure}
    \hspace{0.01\textwidth}
    \begin{subfigure}{0.48\textwidth}
        \includegraphics[width=\textwidth]{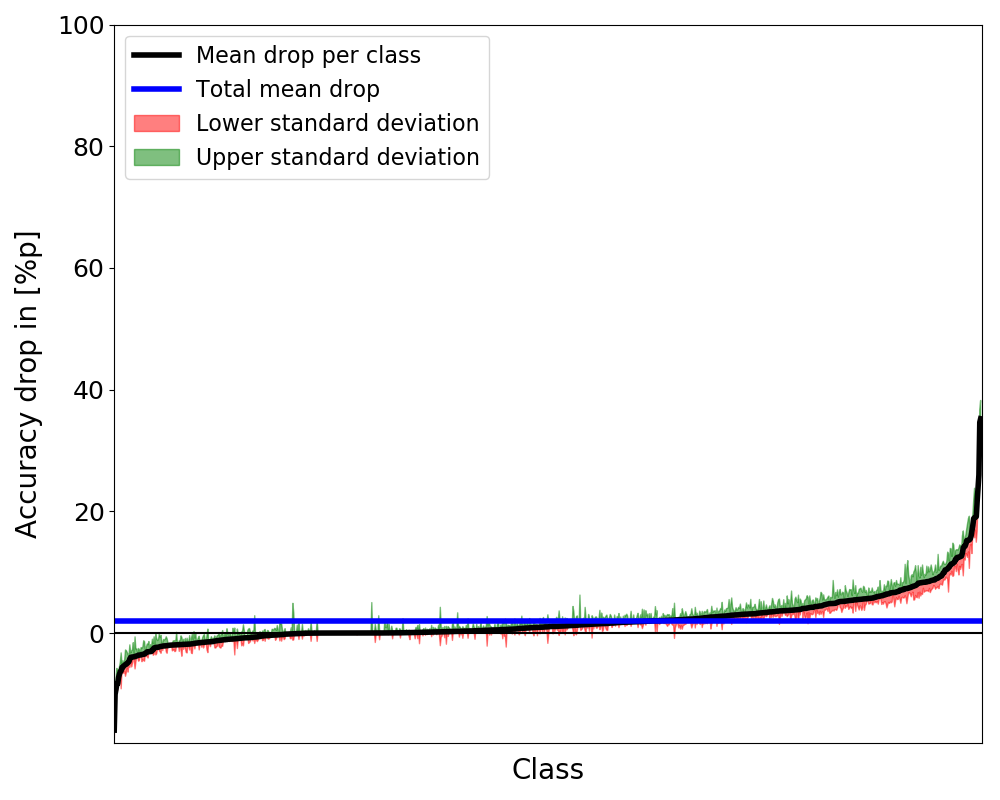}
    \end{subfigure}
    \hspace{0.01\textwidth}
    \begin{subfigure}{0.48\textwidth}
        \includegraphics[width=\textwidth]{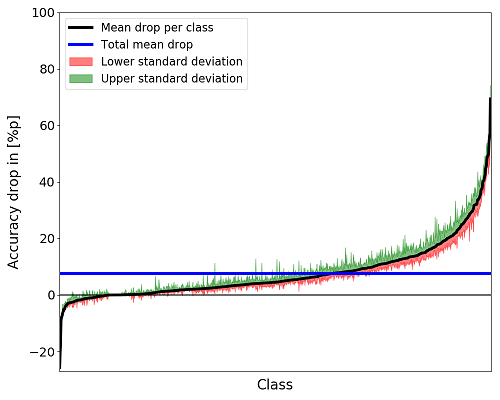}
    \end{subfigure}
    \caption{Top-5 class-specific accuracy drop after ablation of 10\% (left) and 25\% (right) of filters in layer 46 (top) and 49 (bottom)}
    \label{fig:vgg_acc_class_l46l49}
    \vspace*{-10pt}
\end{figure}

Similar to the MLP study, we checked whether the importance of the layers for the overall classification performance shows class-specific variations. We found that, despite the general class-average trend (c.f. Figure \ref{fig:vgg_acc_top5}, black line), some layers are much more important for specific classes than for others. Figure \ref{fig:vgg_acc_top5_class} shows the class-specific drop in top-5 accuracy averaged over all ablations for 5 example classes in addition to the average drop in top-5 accuracy as Figure \ref{fig:vgg_acc_top5}. 
In the case of the 10\% ablations, class 50 shows a much higher drop in accuracy relative to the other classes after ablations in layer 7 and 20 and at the same time a lower drop in accuracy relative to the other classes after ablations in layer 33. Additionally, the drop in accuracy after the 25\% ablation in layer 14 and 17 is much stronger and much weaker, respectively, than the average drop. This observation suggests that layers exhibit a certain degree of class-selectivity and therefore have different relative importance for the overall performance depending on the class. Based on this finding we further investigated how this selectivity is distributed across classes, i.e. to what extent a layer represents specific classes more than others. Figure \ref{fig:vgg_acc_class_l46l49} shows two extreme examples for the class-specific drop in top-5 accuracy after ablations of 10\% (left) and 25\% (right) in layer 46 (top) and layer 49 (bottom), respectively. Consistent with the observations of the MLP study, ablations had a negative effect on the classification performance for most classes. For some classes, however, the class-specific top-5 accuracy improved after the ablations. This effect was stronger for smaller ablations and in layers with a comparably small impact on the overall performance, such as layer 49 (c.f. Figure \ref{fig:vgg_acc_class_l46l49}, bottom right). 

\subsection{Recovery Training of the VGG-19}
\begin{figure}[!t]
    \centering
    \begin{subfigure}{0.48\textwidth}
        \includegraphics[width=\textwidth]{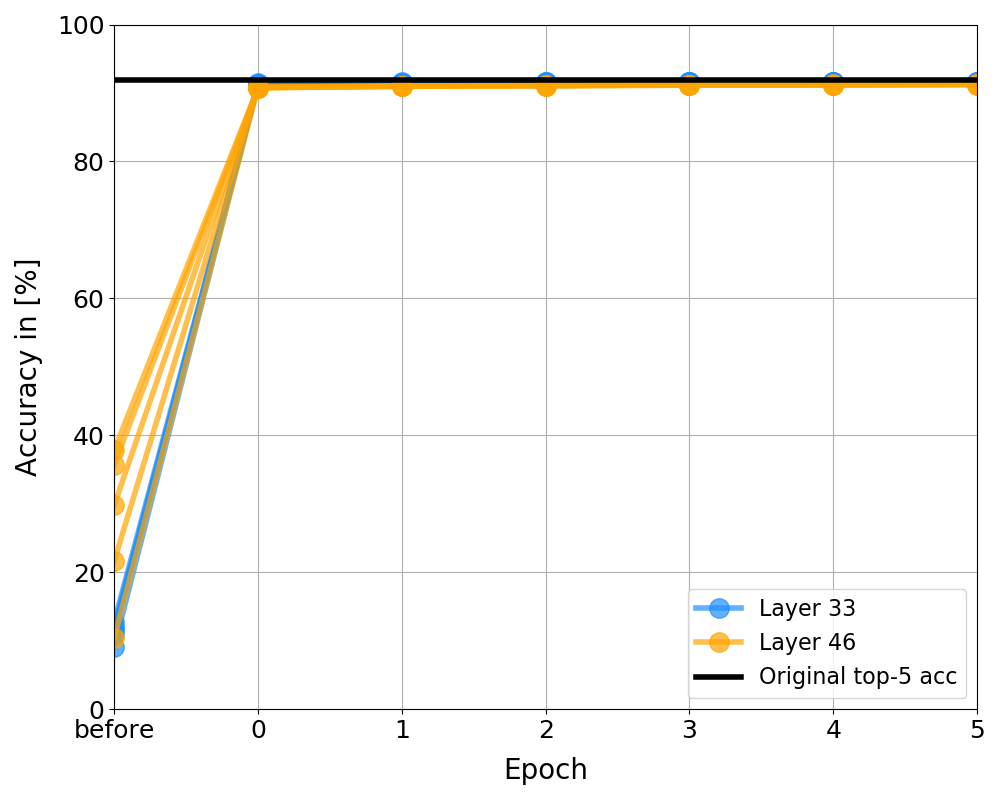}
    \end{subfigure}
    \hspace{0.01\textwidth}
    \begin{subfigure}{0.48\textwidth}
        \includegraphics[width=\textwidth]{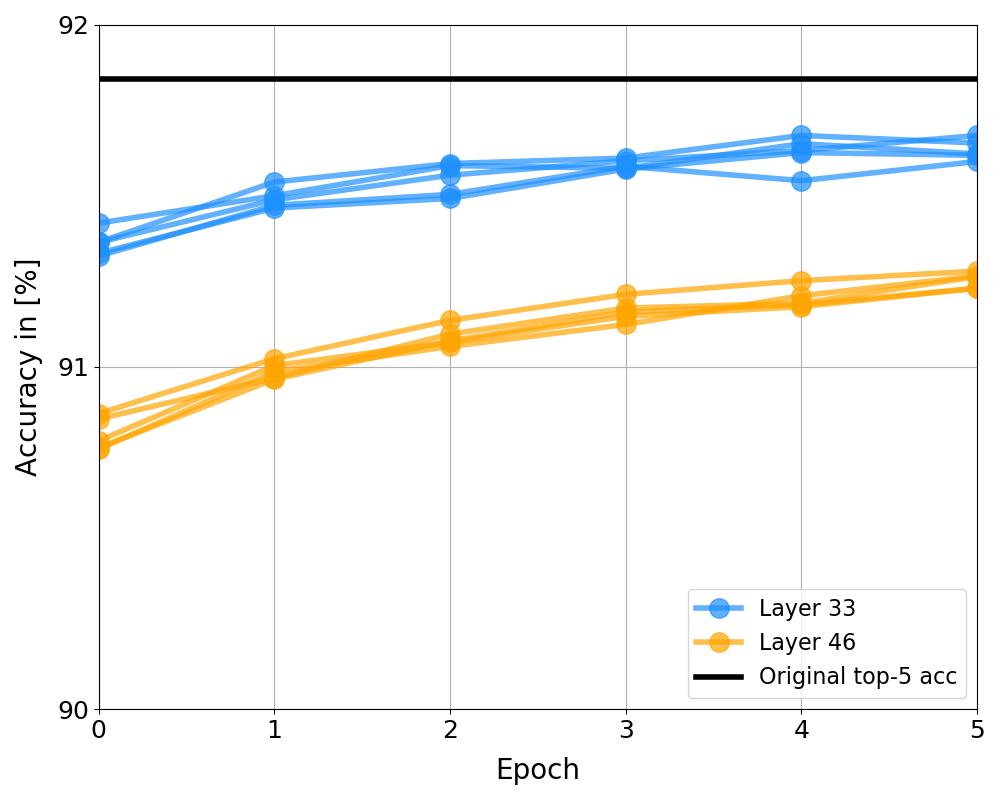}
    \end{subfigure}
    \caption{Recovery process of the top-5 accuracy of 5 instances of the VGG-19 after ablations of 25\% of filters in layers 33 (blue) and 46 (orange)}
    \label{fig:acc5_ex1}
    \vspace*{-10pt}
\end{figure}

Subsequent to the ablations, we aimed to investigate if the negative effects on the classification performance could be recovered and if so, to what extent (c.f. section \ref{ssec:vgg19_rec}). Figure \ref{fig:acc5_ex1} shows the top-5 accuracy after the ablation of 25\% of filters in layer 33 and 46 and 5 epochs of subsequent recovery training in 5 instances of the VGG-19. The results show that the network recovered most of the lost classification ability after a single recovery epoch with a margin of less than 1\%p compared to its original top-5 accuracy (c.f. Figure \ref{fig:acc5_ex1}, left side) with only marginal improvement for the epochs after the first one (c.f. Figure \ref{fig:acc5_ex1}, right side). In general, the original accuracy was never exceeded after the recovery training. However, due to computational cost, recovery training was stopped after 6 epochs, even though the accuracy was still increasing. In the case of layer 46, the extent of the drop in accuracy did not seem to impact the recovery process significantly. Although the top-5 accuracy after the ablations showed a strong variation of up to 30\%p, the network was able to recover the damages regardless of the severity of the initial damage.
\begin{figure}[!t]
    \centering
    \begin{subfigure}{0.48\textwidth}
        \includegraphics[width=\textwidth]{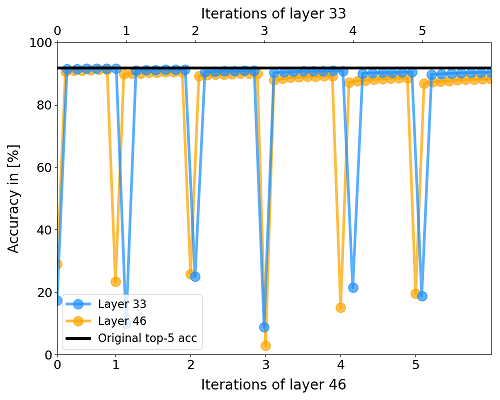}
    \end{subfigure}
    \hspace{0.01\textwidth}
    \begin{subfigure}{0.48\textwidth}
        \includegraphics[width=\textwidth]{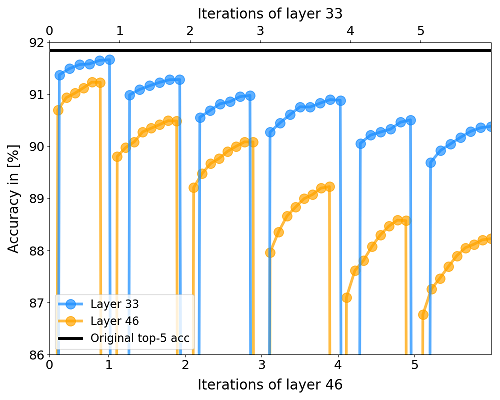}
    \end{subfigure}
    \caption{Iterative ablation of 25\% of filters in layers 33 (blue) and 46 (orange) and subsequent recovery process of the top-5 accuracy of the VGG-19. Note that the filters ablated in each iteration were selected with replacement.}
    \label{fig:acc5_ex2}
    \vspace*{-10pt}
\end{figure}

Figure \ref{fig:acc5_ex2} shows the top-5 accuracy for iteratively performed ablations of 25\% of filters in layer 33 and 46 and subsequent recovery training in a single instance of VGG-19. Note that the filters to be ablated for each iteration were selected with replacement, resulting in a slow and gradual increase of the damage inflicted on the network with each iteration. After the last iteration, $\sim$80\% of the filters were ablated in either layer. Remarkably, the network was able to recover almost completely from the damage caused by the ablations, despite the increasing number of ablated filters. Similar to the first recovery experiment, the performance rapidly increased during the first training epoch for each iteration and only improved marginally for the following epochs. The difference between the recovered top-5 accuracy and the original top-5 accuracy showed a slight increase with the number of iterations. The results suggest that with only $\sim$20\% of filters left in either one of the two most important layers, the network is still able to recover most of the damage and to represent the majority of the necessary information in the remaining network. 

\section{Conclusion and Future Work}
We investigated the effects of single and pairwise unit ablations on the classification performance of a shallow MLP trained on the MNIST dataset. As expected, we found that as a result of ablations, the overall classification performance generally decreased. However, in some cases the class-specific performance for some classes increased despite the overall impairing effect. Furthermore, the ablation of single units revealed their different contributions to the overall classification performance. Some units are universally important for the task, representing features distinct to a majority of the classes in the test set while other units are selectively important for specific classes. Some units turned out to be not important at all. In our future work, we will investigate the origin of the positive effects in detail. The long term goal is to use ablations in a controlled manner to attain a performance which exceeds its backpropagation-trained local optimum.  

Furthermore, we aimed to explain what makes a unit more important than others. We found that the distribution of incoming weights as a structural characteristic of a single unit indicates its importance for the classification task. Specifically, the more a weight distribution changed during training, the more important this unit is for the classification task. This result may prove useful for pruning experiments, as the importance of a unit can be estimated  with significantly reduced computational cost as compared to full functional tests of the network. 

Interestingly, the t-SNE visualizations of the ablation effects revealed that the features represented by single units mostly correspond to the global and local structure inherent to the data set. This suggests that information inherent to the stimuli, with which the network is trained, is mapped and locally represented in specific areas of the network. In a future study, we aim to investigate an underlying organization of this mapping. Possibly, similarities can be discovered to the organized mapping of features inherent to external stimuli, which is found in many areas of the neocortex. 

We further found that the network exhibited robustness against the structural damage caused by ablations, as some features are represented redundantly in different single units. Specifically, pairwise unit ablations revealed effects on the classification performance that exceed the combined effects of the corresponding single unit ablations. Remarkably enough, this observation is true for the negative as well as for the positive effects on the performance. For future work, we aim to draw a parallel to the human brain as a highly complex information processing system exhibiting great robustness to structural damage, which is regarded to be one of its most outstanding features. The research on robustness with respect to structural damage in ANNs has received little attention due to the focus being placed on optimizing network structures with respect to performance, size and speed. In such cases typical for pruning experiments, additional parts of the network representing potentially redundant information are of no special interest. However, we argue that future applications of ANNs in safety-critical domains, such as navigation in autonomous driving, monitoring and control in nuclear power plants or simply privacy protection in the smart home sector, require a new perspective on the importance of robustness of ANNs.

Transferring the principle of ablations to the VGG-19, consistent with the observations for the MLP, we found that ablations had negative as well as positive effects on the classification performance. Specifically, the higher the amount of ablated filters was, the stronger the effect. We further found that the different layers are not equally important for the classification performance. More precisely, the effect of ablations turned out to be significantly stronger for two deeper layers (33 and 46) than for all other layers. At first, this may seem surprising as we expected the upper layers to be the most important layers as they are supposed to represent more general features common to many classes, whereas lower layers are supposed to represent more class-specific features \cite{zeiler2014visualizing}. However, a possible explanation may be that the features represented in these two layers are not redundantly represented in other layers, while the features in the upper layers are. Despite a general trend, the importance of layers showed some class-specific variation. This indicates that the representation of features distinct to specific classes is somewhat localized in the network.

We investigated if and to what extent the inflicted damage to the network could be recovered by subsequent training. We found that irrespective of the location of the ablation and the severity of the inflicted damage, i.e. the magnitude of the drop in accuracy, most of the original classification performance could be recovered after a single epoch of training. Subsequent epochs only improved the performance marginally. However, we found that the larger the amount of ablated filters becomes, the harder it is for the network to recover its original performance. Remarkably though, even after an ablation of $\sim$80\% of filters in the most important layers (33 and 46), the  performance could be recovered up to a difference of less than 4\%p. For the recovery experiments, the ablated filters were selected randomly with replacement. In a future experiment, based on the observation of the positive effects of ablations, we will purposefully choose filters for the ablation and locations to freeze for the subsequent recovery training. The aim is to increase the classification performance to exceed what was originally achieved by means of backpropagation.

\small
\bibliographystyle{ieeetr}
\bibliography{main}

\newpage
\appendix
\renewcommand\thefigure{\thesection.\arabic{figure}}    
\section{Appendix}
\setcounter{figure}{0} 

\begin{figure}[h!]
    \centering
    \begin{subfigure}{0.51\textwidth}
        \includegraphics[width=\textwidth]{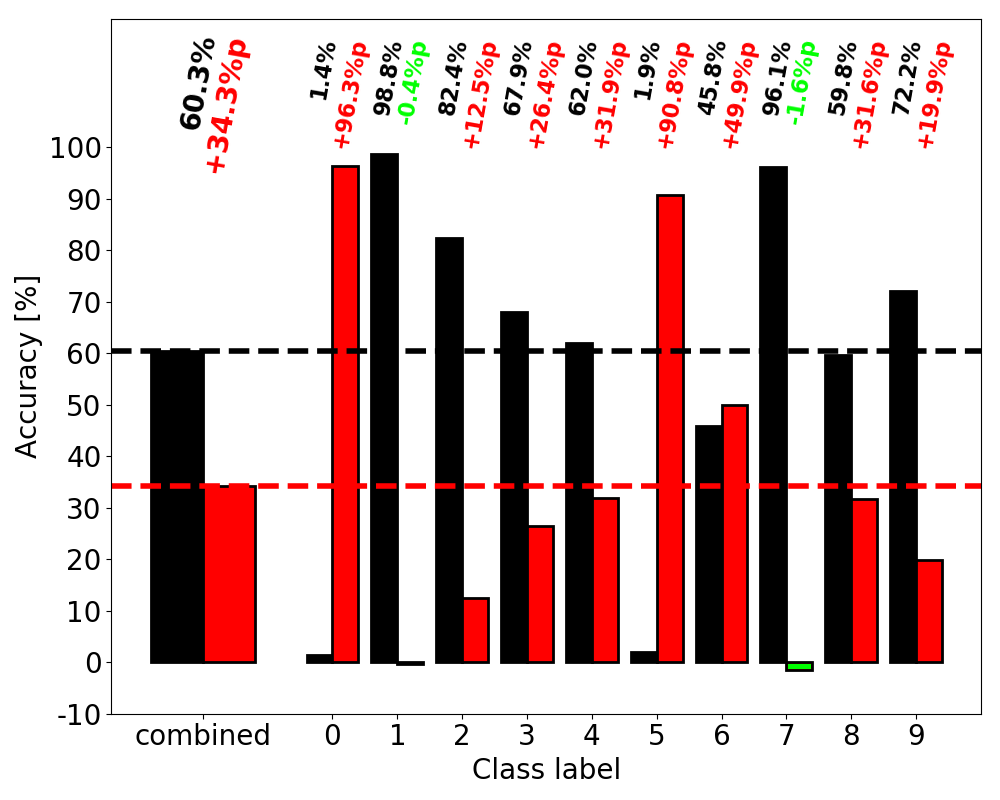}
    \end{subfigure}
    \hspace{0.01\textwidth}
    \begin{subfigure}{0.45\textwidth}
        \includegraphics[width=\textwidth]{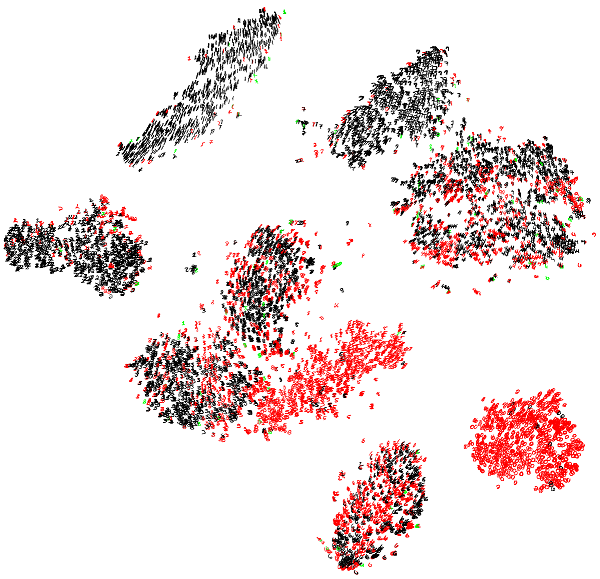}
    \end{subfigure}
    \caption{Overall accuracy, class-specific accuracy and t-SNE visualization of the damaged MLP after the ablation of unit 16 in the first hidden layer. This unit is an example for the representation of features corresponding to many different classes}
    \label{fig:acc-tSNE_ko16}
    \vspace*{-10pt}
\end{figure}
\begin{figure}[h!]
    \centering
    \begin{subfigure}{0.51\textwidth}
        \includegraphics[width=\textwidth]{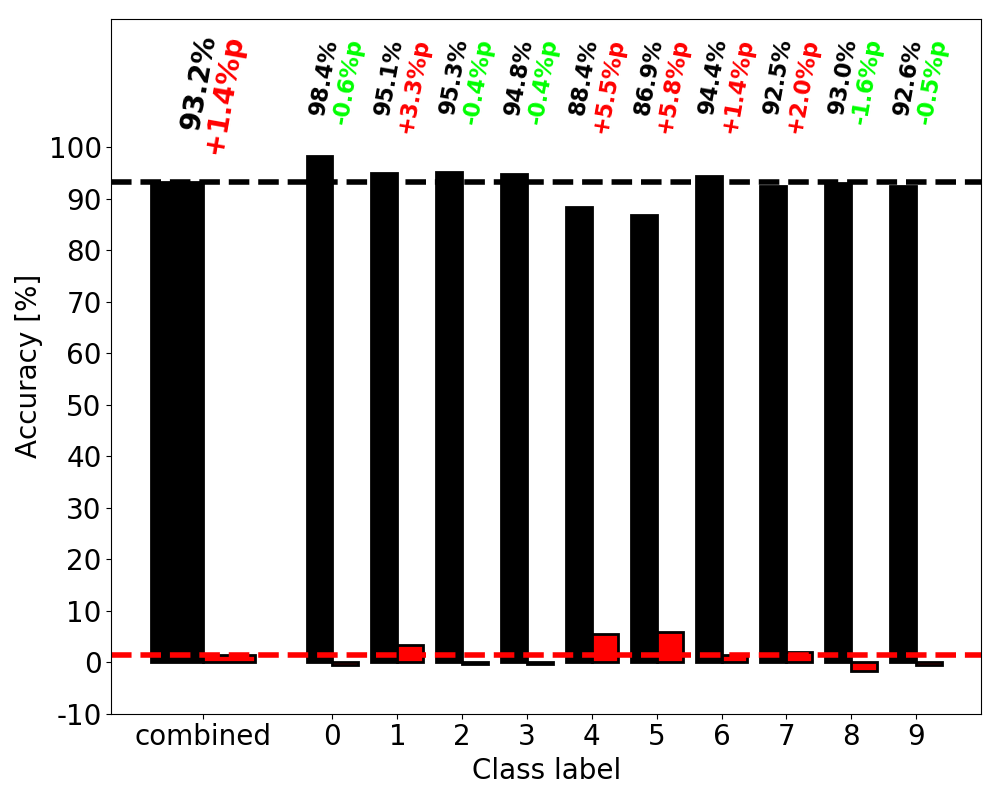}
    \end{subfigure}
    \hspace{0.01\textwidth}
    \begin{subfigure}{0.45\textwidth}
        \includegraphics[width=\textwidth]{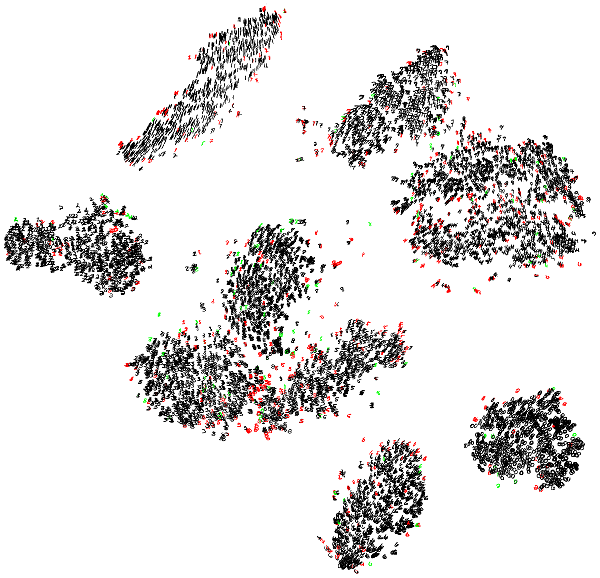}
    \end{subfigure}
    \caption{Overall accuracy, class-specific accuracy and t-SNE visualization of the damaged MLP after the ablation of units 6, 11, 13 and 18 in the first hidden layer. These units do not play a major role in the classification task and would be top candidates for pruning.}
    \label{fig:acc-tSNE_ko6111318}
        \vspace*{-10pt}
\end{figure}
\begin{figure}[h!]
    \centering
    \begin{subfigure}[b]{0.49\textwidth}
        \includegraphics[width=\textwidth]{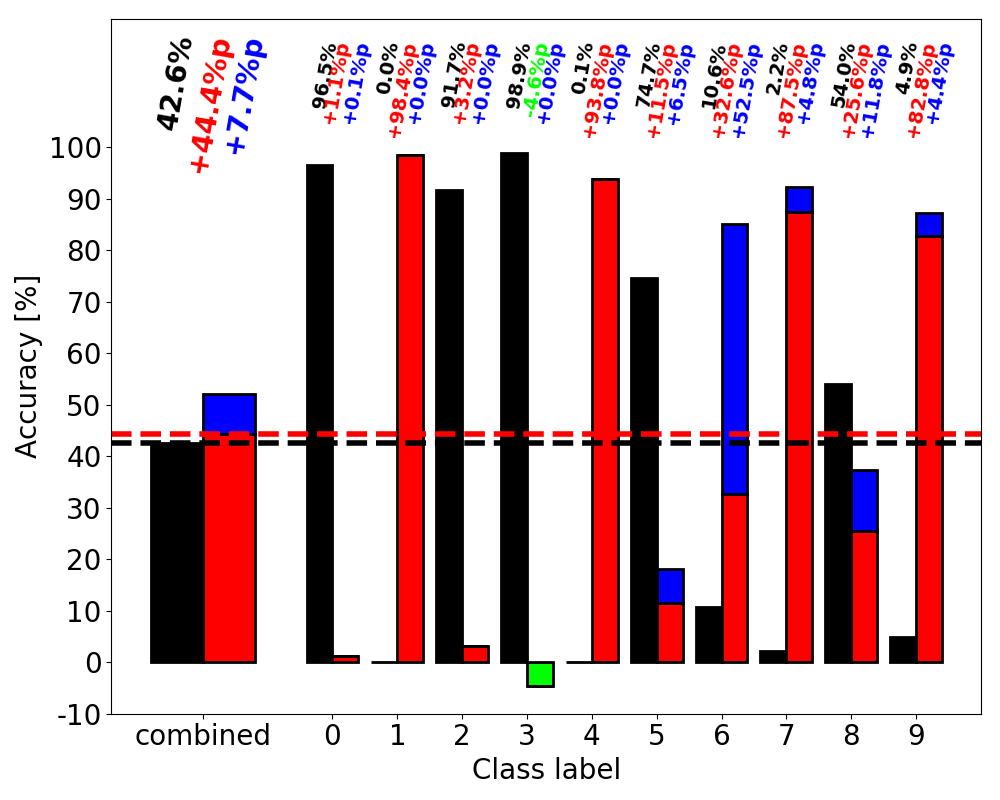}
    \end{subfigure}
    \hspace{0.01\textwidth}
    \begin{subfigure}[b]{0.47\textwidth}
        \includegraphics[width=\textwidth]{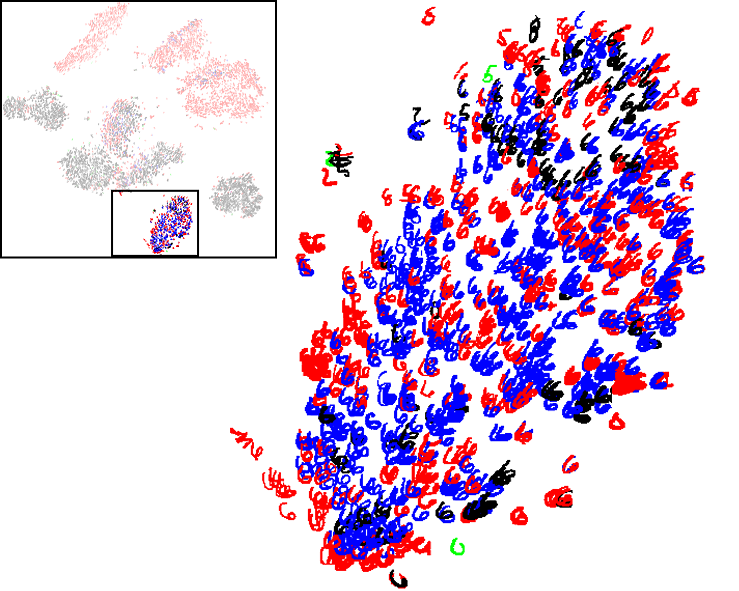}
    \end{subfigure}
    \caption{Overall accuracy, class-specific accuracy and t-SNE visualization of the damaged MLP after the ablation of units 12 and 19 in the first hidden layer. Those units show a strong redundant representation for digits in class six, as a major part of this class is incorrectly classified only after the pairwise unit ablation and not after either single unit ablation.}
    \label{fig:acc-tSNE_ko1219}
    \vspace*{-10pt}
\end{figure}
\begin{figure}[h!]
    \centering
    \begin{subfigure}{0.48\textwidth}
        \includegraphics[width=\textwidth]{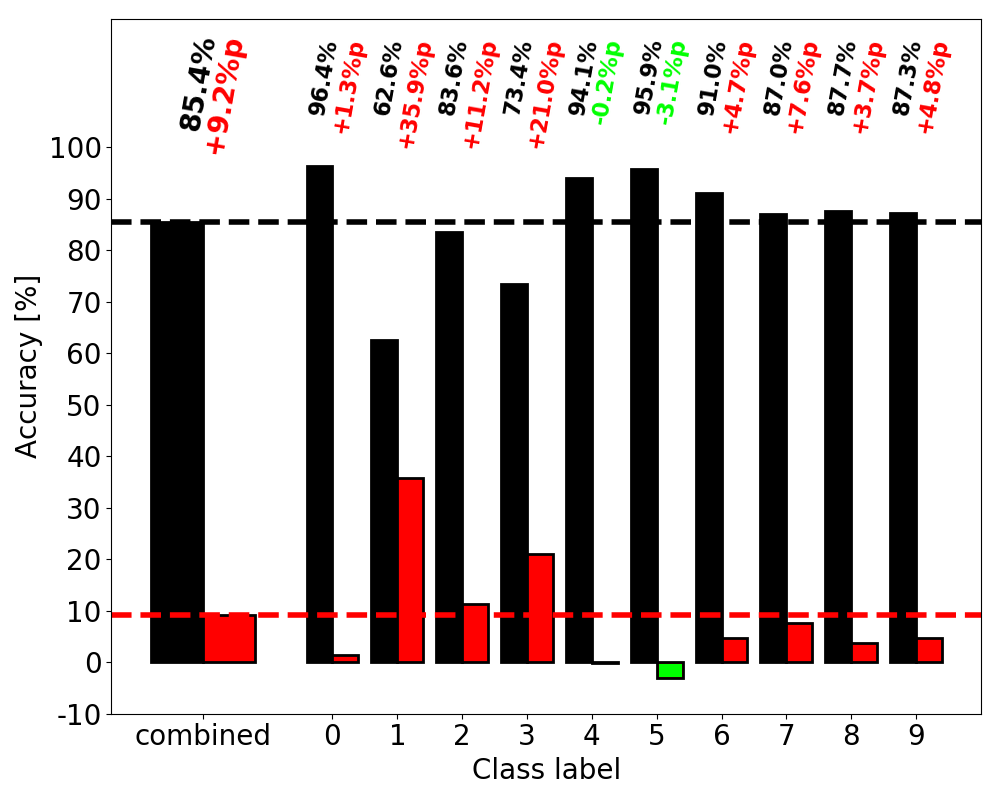}
    \end{subfigure}
    \hspace{0.01\textwidth}
    \begin{subfigure}{0.48\textwidth}
        \includegraphics[width=\textwidth]{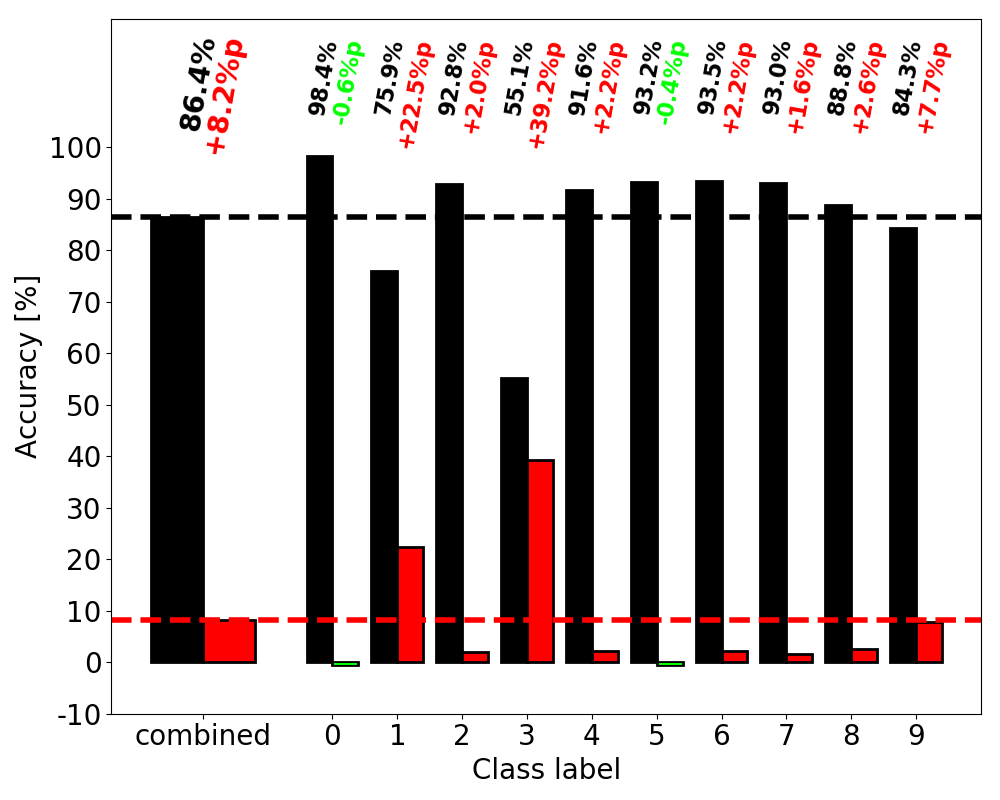}
    \end{subfigure}
    \caption{Overall accuracy and class-specific accuracy after the ablations of unit 5 (left) and 10 (right). Note that the summed positive effect on class five after both single unit ablations is smaller than the positive effect after the pairwise unit ablation (c.f. Figure \ref{fig:acc-tSNE_ko510})}
    \label{fig:acc_ko510}
    \vspace*{-10pt}
\end{figure}

\end{document}